\documentclass[10pt,twocolumn,letterpaper]{article}

\usepackage{cvpr}              

\usepackage[utf8]{inputenc} 
\usepackage[T1]{fontenc}    
\usepackage{url}            
\usepackage{booktabs}       
\usepackage{amsfonts}       
\usepackage{nicefrac}       
\usepackage{microtype}      
\usepackage{xcolor}         

\usepackage{times}
\usepackage{epsfig}
\usepackage{graphicx}
\usepackage{booktabs}
\usepackage{amsmath}
\usepackage{amssymb}
\usepackage{adjustbox}
\usepackage{pifont} 
\usepackage[dvipsnames]{xcolor}
\usepackage[table]{xcolor}
\usepackage{multicol}
\usepackage{multirow}
\usepackage{makecell}
\usepackage{caption}
\usepackage{tcolorbox}
\usepackage{enumitem}
\usepackage{etoc}
\usepackage{colortbl}
\usepackage{pgfplotstable}
\usepackage{placeins}
\usepackage{algorithm}
\usepackage{algpseudocode}

\usepackage[accsupp]{axessibility}  

\newcommand{\cmark}{\textcolor{ForestGreen}{\ding{51}}} 
\newcommand{\xmark}{\textcolor{red}{\ding{55}}}
\newcommand{\tmark}{\textcolor{orange}{$\triangle$}}

\newcommand{\NA}{--}

\newcommand{\upcolor}{ForestGreen}   
\newcommand{\downcolor}{OrangeRed}   

\newcommand{\ty}[1]{\textcolor{orange}{#1}}

\renewcommand{\ty}[1]{\textcolor{black}{#1}}

\newcommand{\datasetname}{HanDyVQA}
\newcommand{\datasetnamenospace}{HanDyVQA}

\definecolor{actColor}{rgb}{1.0, 0.85, 0.85}
\definecolor{proColor}{rgb}{1.0, 0.93, 0.83}
\definecolor{objColor}{rgb}{0.88, 0.88, 0.88}
\definecolor{locColor}{rgb}{0.84, 0.95, 0.86}
\definecolor{staColor}{rgb}{0.83, 0.93, 0.94}
\definecolor{parColor}{rgb}{0.93, 0.88, 0.96}

\definecolor{cvprblue}{rgb}{0.21,0.49,0.74}
\usepackage[pagebackref,breaklinks,colorlinks,allcolors=cvprblue]{hyperref}

\def\paperID{} 
\def\confName{CVPR}
\def\confYear{2026}

\title{HanDyVQA: A Video QA Benchmark for\\Fine-Grained Hand-Object Interaction Dynamics}

\author{
Masatoshi Tateno$^{1,2}$\hspace{1em}
Gido Kato$^{3,2}$\hspace{1em}
Hirokatsu Kataoka$^{2,4}$\hspace{1em}
Yoichi Sato$^{1}$\hspace{1em}
Takuma Yagi$^{2}$\\
$^1$Institute of Industrial Science, The University of Tokyo\\
$^2$National Institute of Advanced Industrial Science and Technology (AIST)\\
$^3$Waseda University
$^4$Visual Geometry Group, University of Oxford\\
{\tt\small{\url{https://masatate.github.io/HanDyVQA-project-page/}}} \\
}

\begin{document}
\twocolumn[{%
\renewcommand\twocolumn[1][]{#1}%
\maketitle
\begin{center}
    \centering
    \captionsetup{type=figure}
    \includegraphics[width=0.95\textwidth]{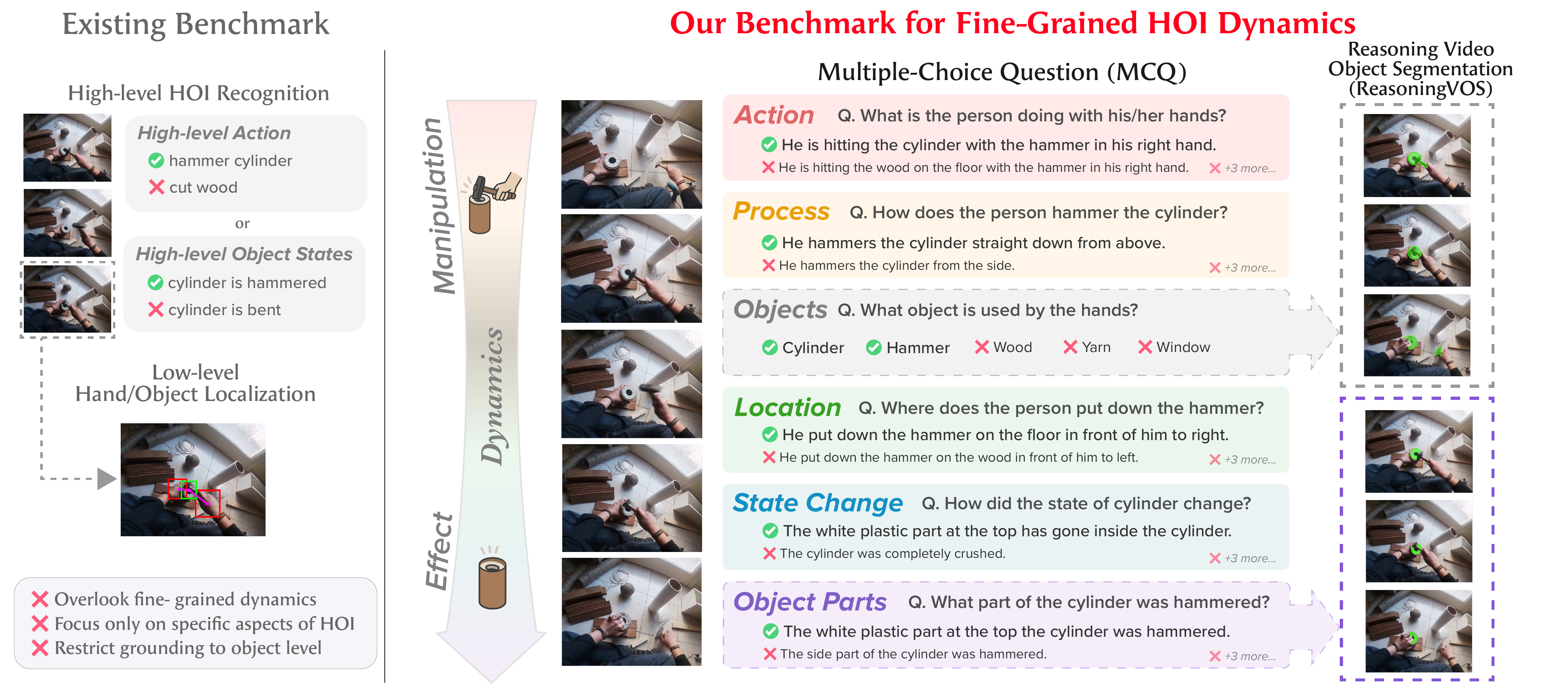}
    \captionof{figure}{Overview of \datasetname~Dataset. \datasetname~evaluates fine-grained hand–object interaction (HOI) dynamics for both manipulation and effect aspects through MCQ and ReasoningVOS tasks.
    MCQ only shows a subset of answer candidates.}
    \label{fig:teaser}
\end{center}%
}]



\begin{abstract}

Hand–object interaction (HOI)
involves dynamics where human manipulations produce
spatio-temporal effects on objects.
%
However, existing semantic HOI benchmarks focus on either manipulation or effects at a coarse level, lacking fine-grained spatio-temporal reasoning to capture HOI dynamics.
We introduce HanDyVQA, a fine-grained video
QA benchmark that comprehensively covers both the manipulation and effect aspects of HOI. \datasetname~comprises six complementary question types (Action, Process, Objects, Location, State Change, and Object Parts), totaling 11.1K multiple-choice QA pairs.
Collected QA pairs require recognizing manipulation styles, hand/object motions, and part-level state changes.
HanDyVQA also includes 10.3K segmentation masks for Objects and Object Parts, enabling the evaluation of object/part-level reasoning in video object segmentation.
%
We evaluated recent video foundation models on our benchmark and found that even the best, Gemini-2.5-Pro, achieved only 73\% accuracy, well below human performance (97\%). 
Further analysis shows the remaining challenges in spatial relationships, motion, and part-level geometric understanding.
%
We also found that incorporating explicit HOI cues into visual features improves performance, providing insights for future HOI-aware models.
\end{abstract}

\begin{table*}[t]
\adjustbox{max width=\textwidth}{
\begin{tabular}{lcccccccccc}
\hline
  & \multirow{2}{*}{\textbf{Source}} & \multirow{2}{*}{\textbf{Task Type}} & \multicolumn{2}{c}{\textbf{Fine-grained Dynamics}} & \multirow{2}{*}{\textbf{\# Domain}} & \textbf{Full Human} &\multicolumn{2}{c}{\textbf{Answer Type}} & \multirow{2}{*}{\textbf{\# Question}}  & \multirow{2}{*}{\textbf{Avg. Duration}} \\ \cline{4-5} \cline{8-9}
  &  &  & Manipulation & Effect & & \textbf{QA Check} & Semantic & Grounding &  \\ \hline
EgoTaskQA~\cite{jia2022egotaskqa}  & LEMMA & Causal / Intents / Belief & \xmark  & \xmark & 1\textasciitilde2 & \xmark & OP & \xmark & 40K  &  25~s \\
EgoThink~\cite{cheng2024egothink} & Ego4D & Reasoning / Forecasting / Planning & \xmark & \xmark & 26 & \cmark & OP & \xmark & 700 & Single frame \\
HOI-QA~\cite{bansal2024hoi} & EK/Ego4D & Hand and Object Location Referral  & \xmark & \xmark & 40 & \xmark & OP & BBox & 3.9M & Single frame \\
EgoHOIBench~\cite{xu2024egonce++} & Ego4D & Action / Objects  & \xmark & \xmark & 74 & \xmark & MC & \xmark & 30K & 1~s \\
AMB~\cite{goletto2024amego} & EK & Long-Term Object Interactions & \xmark & \xmark & 1 & \xmark & MC & \xmark & 21K &  20~m \\
OSCAR~\cite{nguyen2024oscar} & EK/Ego4D & Object State Change Understanding  & \xmark & \cmark & 52 & \xmark & OP \& MC & \xmark & 400K & Unavailable \\
HD-EPIC~\cite{perrett2025hd} & HD-EPIC & Fine-grained Video / 3D Understanding & \cmark & \tmark\textsuperscript{*1}  & 1 & \xmark & MC & \xmark & 26K &  Variable \\
\datasetname (Ours) & Ego4D & Dynamics of Manipulation and Effect & \cmark & \cmark & 112 & \cmark & MC & Seg & 11K &  5~s \\ \hline
\end{tabular}
}
\caption{Comparison against HOI-related QA datasets: We calculate number of video domains based on Ego4D~\cite{song2023ego4d} scenario taxonomy. MC stands for multiple-choice question-answering and OP represents open-ended question-answering. BBox indicates bounding box, and Seg refers to segmentation.  Our dataset covers both manipulation and effect dynamics in HOI, with segmentation masks in diverse real-world domains.  All QA pairs are verified by human. \textsuperscript{*1} \footnotesize{HD-EPIC includes only Location questions for fine-grained effect understanding, excluding categories such as State Change and Object Parts.}}
\label{tab:dataset_comparison}
\end{table*}

\section{Introduction}
\label{sec:intro}

Hand-Object Interaction (HOI) is inherently a dynamic process~\cite{gibson2014ecological}. 
To accomplish tasks with precision, humans skillfully coordinate their hand movements to manipulate objects or tools, thereby inducing the desired effect on the target objects.
Accurately recognizing the spatiotemporal dynamics in hand-object interactions opens up various applications, such as worker assistance~\cite{flaborea2024prego}, dexterous manipulation in robots~\cite{shaw2023videodex}, and motor function analysis~\cite{tsai2023recognizing}.

While there has been a surge in semantic HOI understanding methods and benchmarks in recent years, most existing studies focus on high-level understanding of either human actions --- including action recognition~\cite{damen2020epic,kwon2021h2o,grauman2022ego4d,xu2024egonce++}, long-form actions~\cite{mangalam2023egoschema}, and procedural steps~\cite{sener2022assembly101,song2023ego4d,yagi2025finebio} --- or object states~\cite{tateno2025learning, xue2024learning, nguyen2024oscar}.
However, these benchmarks isolate a single aspect of HOI, overlooking its inherently dynamic nature that arises through the progression from human manipulation to the resulting effects (e.g., how a person hammers a cylinder and how and where the object’s state changes).




We propose \datasetname\  (\textbf{Hand} \textbf{Dy}namics \textbf{V}ideo \textbf{QA}), a video question-answering benchmark designed to evaluate
the dynamic process of HOI, spanning from manipulation to effect (see Figure~\ref{fig:teaser}).
\datasetname\ requires understanding not only the actions and objects involved, but also their processes, effects, and component-level changes. 
The benchmark provides six types of multiple-choice question answering (MCQ) tasks, totaling 11.1K QA pairs, that cover both the manipulation (\textbf{Action}, \textbf{Process}, \textbf{Objects}) and effect (\textbf{Location}, \textbf{State Change}, \textbf{Object Parts}) aspects of HOI.
\textbf{Objects} and \textbf{Object Parts} further include reasoning video object segmentation (ReasoningVOS) tasks, totaling 10.3K frames, which requires implicit pixel-level reasoning guided by the question, unlike ReferringVOS~\cite{khoreva2018video} based on explicit textual grounding.
We built the benchmark on short video clips extracted from the Ego4D~\cite{grauman2022ego4d} dataset, enabling the creation of annotations across diverse HOI scenarios in real-world settings. Compared to other HOI-related VQA datasets, our benchmark spans a wide range of activity domains (See Table \ref{tab:dataset_comparison}).
Notably, all question–answer pairs are manually verified by humans for both quality and difficulty.

We evaluate existing video-language models to quantify how well they capture the dynamics of HOI. 
Our results show that even the latest foundation models struggle across all categories. Even the best-performing model, Gemini-2.5-Pro \cite{comanici2025gemini}, achieves only around 68--79\% accuracy in MCQ while human achieves more than 95\% for all categories.
Ablation studies suggest that increasing the number of input frames and resolution effectively enhances overall performance.
Nevertheless, an error-specific analysis indicates that the reduction in motion-related errors remains limited.
Furthermore, the ReasoningVOS results reveal that current models often overlook multiple manipulated objects or over-segment queried object parts.
Consequently, the score on \textbf{Object Parts} is substantially lower than typical Referring / Reasoning VOS benchmarks, posing a new challenge for component-level grounding.

Moreover, as a baseline study for understanding HOI dynamics, we investigate whether explicitly incorporating (i) hand pose, (ii) object tracking, and (iii) object features can enhance the performance.
The results reveal that each component helps reduce different error types, improving performance across categories.
This highlights the importance of video encoders that explicitly model local hand–object information and their spatio-temporal dynamics.

Our contribution is as follows:
(a) We introduce \datasetnamenospace, a new comprehensive dataset for understanding fine-grained dynamics in HOIs.
(b) We conduct an in-depth analysis of how latest video-language models struggle to capture spatiotemporal dynamics and pixel-level reasoning in HOI.
(c) We show fine-tuning models with additional hand and object information can enhance the performance, showing the necessity of modeling fine-grained temporal evolution of hands, objects and their components. 

\section{Related Work}

\paragraph{Hand-object interaction recognition benchmarks.}
Various HOI recognition benchmarks have been proposed with focuses on (i) low-level localization and (ii) high-level actions or object states. For the former, benchmarks have focused on detecting hands and objects-in-contact~\cite{shan2020understanding}, estimating 3D hand and object poses~\cite{hampali2020honnotate, chao2021dexycb}, reconstructing mesh representations~\cite{swamy2023showme}, and object tracking~\cite{banerjee2025hot3d, goletto2024amego}.
For the latter, benchmarks built on egocentric datasets such as EPIC-KITCHENS~\cite{damen2020epic} and Ego4D~\cite{grauman2022ego4d} cover high-level HOI understanding, including action recognition~\cite{damen2020epic,kwon2021h2o,grauman2022ego4d,xu2024egonce++, bansal2024hoi, perrett2025hd}, action forecasting and planning~\cite{cheng2024egothink}, long-form actions~\cite{mangalam2023egoschema}, and procedural steps~\cite{sener2022assembly101,song2023ego4d,yagi2025finebio}. Object state recognition benchmark includes the temporal localization of state (change)~\cite{song2023ego4d, xue2024learning,tateno2025learning, zameni2025moscato} and state captioning~\cite{nguyen2024oscar}. 
While these works address aspects of HOI understanding, they remain limited to either manipulation or effects, lacking comprehensive evaluation of fine-grained spatio-temporal dynamics.


\paragraph{Referring/Reasoning video object segmentation.}
Referring Video Object Segmentation (RVOS)~\cite{khoreva2018video, seo2020urvos,ding2023mevis} aims to segment the target object in a video given a natural language expression, while Reasoning Video Object Segmentation (ReasoningVOS)~\cite{bai2024one, yan2024visa} introduces implicit textual queries that require complex reasoning over world knowledge and video contents.
EgoMask~\cite{liang2025fine} introduced an RVOS benchmark for egocentric video, which is challenging due to fast camera motion and dense object scenes. HOI-QA~\cite{bansal2024hoi} proposed hands and objects localization task from referring expressions in egocentric images.
However, these benchmarks remain limited to object-level grounding with direct expressions.
Our benchmark advances this by introducing object- and part-level ReasoningVOS that require reasoning over dynamic hand–object relationships and structural changes in egocentric videos.


\paragraph{MLLMs and Benchmarks for video understanding.}
Recent video-language models integrate visual encoders pretrained on large-scale image/video-text pairs~\cite{radford2021learning, zhai2023sigmoid, wang2024internvideo2} with LLMs, forming multimodal LLMs (MLLMs) that demonstrate strong generalization across diverse video understanding tasks~\cite{bansal2024hoi, zhang2024video, cheng2024videollama, ye2024mplug, wang2024qwen2}.
In parallel, various benchmarks have been introduced to evaluate understanding of general video~\cite{li2024mvbench, xiao2021next, fang2024mmbench}, long-video~\cite{wu2024longvideobench,wang2025lvbench,zhou2025mlvu,mangalam2023egoschema}, egocentric video~\cite{damen2020epic, grauman2022ego4d}, and event/moment localization and retrieval~\cite{xu2016msr, lei2021detecting}.
Recent progress has largely focused on scaling model size and training data to cover diverse tasks.
However, most MLLMs rely on simple frame-based architectures and rarely model local entities or spatio-temporal dynamics, such as hand poses, manipulated objects, and fine-grained state or structural changes.
Existing benchmarks also do not evaluate these aspects.
\datasetname~addresses this gap by introducing a new challenge that promotes visual encoders capable of reasoning about complex hand–object interactions in dynamic scenes.

\section{\datasetname~Benchmark}
\label{sec:handyvqa}

Our goal is to build a systematic benchmark that evaluates models’ ability to recognize the spatiotemporal dynamics in HOI manipulation and effects.
To this end, we introduce two tasks in our benchmark: (1) Multiple-Choice Question (MCQ) and (2) Reasoning Video Object Segmentation (ReasoningVOS). Given a video and a question, the goal of the MCQ task is to select the correct answer(s) from a set of options, while the ReasoningVOS task requires predicting the segmentation masks corresponding to the correct answer. We define six question categories: Action, Process, Objects, Location, State Change, and Object Parts. MCQ samples are provided for all question types, whereas ReasoningVOS samples are provided only for Objects and Parts questions.

We adopt the MCQ format over open-ended questions
to reduce ambiguity in evaluation and probe fine-grained HOI understanding through challenging distractors.
In this section, we describe our data collection process (\cref{subsec:qa_collection}) and its analysis (\cref{subsec:statistics}).

\subsection{QA collection}
\label{subsec:qa_collection}
We developed a collaborative framework that uses LLMs to propose initial QA candidates that are carefully refined and verified by humans to ensure quality and diversity.

\paragraph{Data curation.}
We build our benchmark on Ego4D~\cite{grauman2022ego4d} as it includes unscripted and realistic hand-object interactions across variety of scenarios and recording locations.
We curate HOI video clips using narration annotations with timestamps by prompting LLMs to identify whether any object is being manipulated in each action.
After curation, we sample 2,000 narrations per category that contain relevant verbs (primary action conducted in the clip) or contextual information suitable for each question type. Especially for \textbf{Object Parts} question, we use LLMs to filter the narration that implies partial state change of objects. For each narration, we use a 5-second video segment centered around its timestamp, spanning 2.5 seconds before and after the narration.
See supplementary for details.

\paragraph{Question candidate generation.}
We automatically generate candidate questions from narrations using the following templates.
\textbf{Action:} ``What is the person doing with his/her hands?''
\textbf{Process:} ``How does the person \texttt{[verb]} \texttt{[object]}?''
\textbf{Objects:} ``What object is used by the hands?''
\textbf{Location:} ``Where does the person \texttt{[verb]}''
\textbf{State Change:} ``How did the state of \texttt{[object]} change?''
\textbf{Object Parts:} ``What part of \texttt{[object]} is \texttt{[effect]}?''

Verbs and objects are extracted from the narration and inserted to the corresponding placeholders, \texttt{[verb]} and \texttt{[object]}. For \textbf{Object Parts} questions, we ask LLMs to infer the plausible objects and effects to be inserted.
While most questions adhere to these templates, minor revisions are introduced to fit the grammar and context during the human refinement stage.

\paragraph{QA refinement by humans.}
Given the generated questions, annotators verify their validity and revise or reject any that do not match the actual content. Then, they provide a correct answer for each question---listing all plausible objects in the {\bf Objects} category where multiple answers may exist---while ensuring that each answer contains enough details.
Next, distractor candidates are generated by LLMs from the question-answer pair.
Annotators then refine these candidates by removing overlaps, improving plausibility, and adding more challenging distractors when necessary.
Overall, annotators ensure that all questions, answers, and choices are accurate, sufficiently confusing, and solvable by humans.
Examples of challenging questions are shown in \cref{fig:mcq_qualitative}.

\paragraph{Mask annotation by humans.}
For the \textbf{Objects} and \textbf{Object Parts} questions, annotators sample about three representative frames from each video, distributing them as evenly as possible across the video, where the target regions were clearly visible, and annotated the corresponding regions.

\begin{figure}[t]
    \centering
    \includegraphics[width=\linewidth]{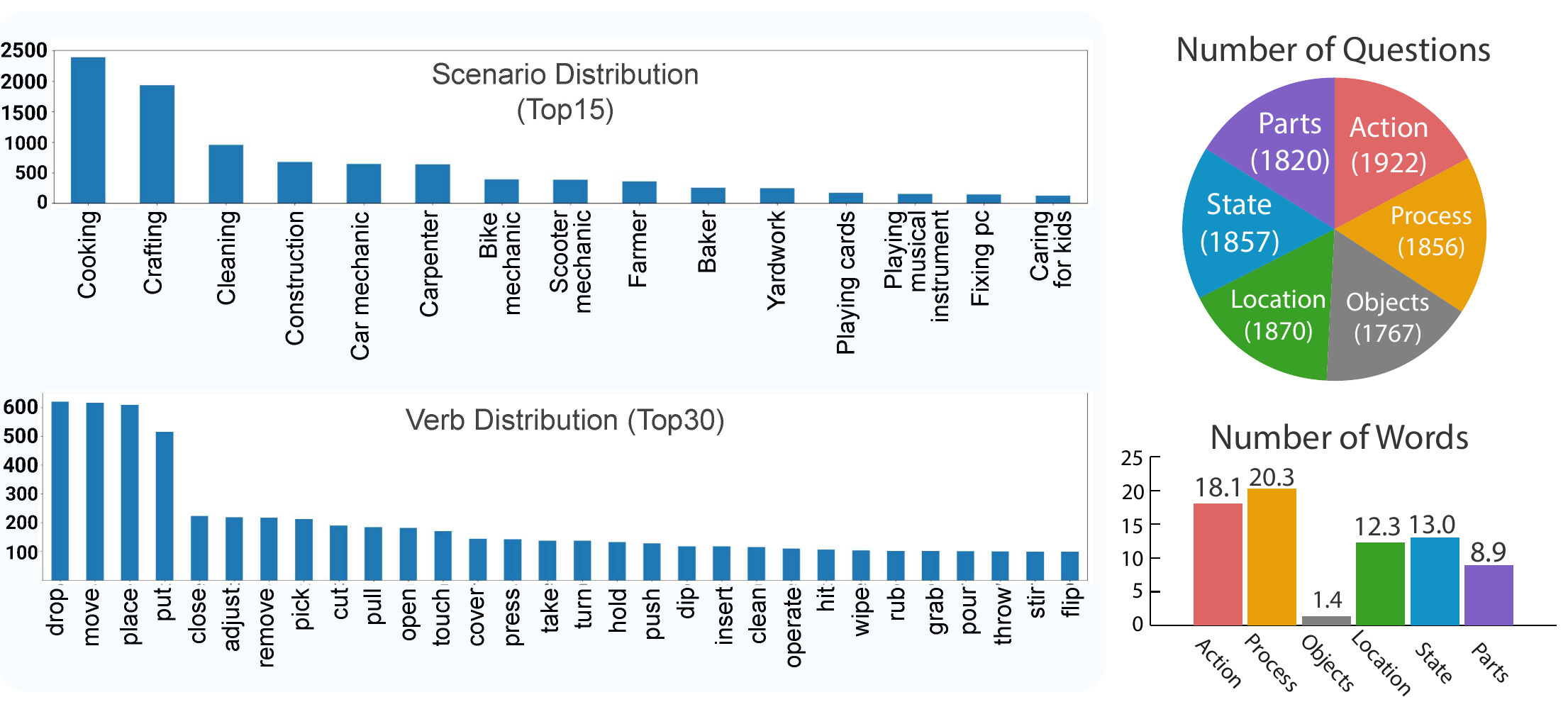}
    \caption{Scenario distribution of \datasetname~dataset (left), number of questions per question type (top right), and average number of words per option (bottom right).}
    \label{fig:diversity}
\end{figure}

\begin{table}[t]
    \raggedright
    \resizebox{\linewidth}{!}{%
    \begin{tabular}{lcccc}
         \toprule
            &  \makecell{\textbf{\#Frames}} 
            & \makecell{\textbf{Avg. Frames} \\ \textbf{per Video}} 
            & \makecell{\textbf{Avg. Centroid}  \\ \textbf{Shift} (px) \footnotesize{*1}} 
            & \makecell{\textbf{Avg. IoU b/w} \\ \textbf{Adjacent Frames} } \\
            \midrule
            \textbf{Objects} & 5053 & 2.88 & 93.70 & 0.17 \\
            \textbf{Parts} & 5216 & 2.89 & 88.35 & 0.08 \\
            \bottomrule
    \end{tabular}
    }
    \scriptsize{$*1$ GT-mask centroid displacement between adjacent frames, computed at 480×854.}
    \caption{Statistics of mask annotations for \textbf{Objects} and \textbf{Parts}.}
    \label{tab:spatiotemporal_stats}
\end{table}

\subsection{Dataset statistics}
\label{subsec:statistics}
After all human verification, we obtain a total of 11,092 QA pairs.
Each question has one correct answer out of five options.
For the \textbf{Objects} category, multiple objects are often manipulated within a 5-second clip, leading to more than five options and multiple correct answers on average (5.7 options and 1.6 answers).
For options, {\bf Action} and {\bf Process} exhibit longer descriptions than other categories as they provide detailed explanations of the hand side, the objects involved, and the fine-grained hand motions. (see \cref{fig:diversity})

\paragraph{Diversity in HOI scenarios.}
As shown in \cref{fig:diversity}, our dataset covers a wide range of video scenarios, including cooking, gardening, cars, and more. We also cover diverse verbs in the narration annotations, requiring the models to understand various actions and their underlying interactions.

\paragraph{Distribution of mask annotation.}
Table \ref{tab:spatiotemporal_stats} shows the number of annotated frames and the relative movement/spatial overlap between them. 
Due to moving objects and cameras in egocentric videos, the segmentation masks shift dynamically over time and space, making them challenging to predict.
See supplementary for further analysis.

\paragraph{Splits.}
We divided the videos into training, validation, and test sets in a 10 : 5 : 85 ratio, yielding 1.1 K, 0.5 K, and 9.4 K questions, respectively.
Only a small portion was set aside as training/validation sets for instruction tuning, allowing models to learn the required output format while placing greater emphasis on evaluation rather than model training.

\paragraph{Human evaluation.}
We conducted a human evaluation on the MCQ task with three professional annotators. As shown at the bottom of \cref{tab:comparison}, humans achieved over 95\% across all question types, confirming that the QA pairs are solvable.
\begin{table*}[ht]
    \raggedright
    \resizebox{\linewidth}{!}{%
    \small
    \begin{tabular}{lccc|cccccc|c}
        \toprule
        \textbf{Models} & \multirow{2}{*}{\textbf{Visual Backbone}} & \multirow{2}{*}{\textbf{Resolution}} & \multirow{2}{*}{\textbf{LLM}} & \cellcolor{actColor} \textbf{Action} & \cellcolor{proColor} \textbf{Process} & \cellcolor{objColor} \textbf{Objects}  & \cellcolor{locColor} \textbf{Location} & \cellcolor{staColor} \textbf{State} & \cellcolor{parColor} \textbf{Parts} & \textbf{Avg.}\textsuperscript{*3} \\
         (Zero-shot) & & & & (Acc) & (Acc) & (AP) & (Acc) & (Acc) & (Acc) & (Acc)  \\
        \midrule
        \midrule
        Random & \NA & \NA & \NA & 19.3 & 18.9 & 28.6 & 20.4 & 19.8 & 19.4 & 19.5 \\

        \midrule
        \multicolumn{3}{l}{\textit{\textcolor{gray}{Text only models}}} &&&&&&&&\\
        GPT-4o\textsuperscript{*1} (text)~\cite{hurst2024gpt} & \NA & \NA & GPT-4o  & 36.6 & 50.9 & 34.3 & 34.1 & 39.5 & 45.5 & 41.3 \\
        \midrule
        \multicolumn{3}{l}{\textit{\textcolor{gray}{Open-source dual-encoder video-language models}}} &&&&&&&&\\
        LaViLa (TSF-L)~\cite{zhao2023learning} & TimeSformer & 224x224 & \NA & 61.6 & 40.1 & 68.5 & 36.9 & 38.9 & 35.6 & 42.6 \\

        InternVideo2-Stage2~\cite{wang2024internvideo2} & Original & 224x224 & \NA  & 41.1 & 30.2 & 37.0 & 29.7 & 34.9 & 30.5 & 33.3 \\

        \multicolumn{3}{l}{\textit{\textcolor{gray}{Open source video-language models w/ integrated LLMs}}} &&&&&&&&\\
        VideoLLaMA2.1-7B~\cite{cheng2024videollama} & SigLIP & 384x384 & Qwen2 & 41.4 & 47.3 & 52.9 & 34.6 & 47.0 & 40.5 & 42.1 \\
        
        LLaVa-Video-7B~\cite{lin2024video} & SigLIP & 384x384 & LLaVa-7B & 56.9 & 53.7 & 60.4 & 50.5 & 58.5 & 54.6 & 54.8 \\
        
        mPLUG-Owl3-8B~\cite{ye2024mplug} & SigLIP & 384x384 & Qwen2 & 52.1 & 53.1 & 61.0 & 45.8 & 54.7 & 48.6 & 50.9 \\
        

        Qwen2.5-VL-7B~\cite{wang2024qwen2} & Original & 384x384 & Qwen2.5 & 60.8 & 54.9 & 53.9 & 47.9 & 56.7 & 48.6 & 53.8 \\


        Qwen2.5-VL-72B~\cite{wang2024qwen2} & Original & 480x854 & Qwen2.5 & 78.0 & 73.4 & 75.2 & 63.2 & 72.2 & 62.5 & 69.9 \\
        \midrule
        \multicolumn{3}{l}{\textit{\textcolor{gray}{Proprietary vision and language models w/ integrated LLMs}}} &&&&&&&&\\
        GPT-4o\textsuperscript{*1} (vision)~\cite{hurst2024gpt} & Original & 480x854 & GPT-4o & 61.3 & 64.4 & 64.1 & 51.5 & 59.0 & 58.5 & 58.9 \\
        Gemini-2.5-Pro~\cite{comanici2025gemini} & Original & 480x854 & Gemini-2.5-Pro & 79.1 & 73.3 & 78.8 & 67.6 & 73.9 & 69.3 & 72.6 \\
        \midrule
        \textbf{Human}\textsuperscript{*2} & \NA & \NA & \NA & 98.6 & 95.9 & 96.0 & 96.6 & 95.3 & 96.9 & 96.6 \\
        \bottomrule
    \end{tabular}
    }
    \scriptsize{$*1$ GPT-4o text/vision refused to answer some questions, providing valid answers to around 87\% and 79\% of total questions. }
    \scriptsize{$*2$ We report average score of three human participants. }
    \scriptsize{$*3$ We average scores of all categories except Objects due to different metrics.}
    \caption{Comparison of zero-shot performance of different models across various question types. We report numbers from valid responses.}
    \label{tab:comparison}
\end{table*}

\section{Experiments}
\label{sec:experiments}





To reveal the challenges in recognizing the dynamic aspects of HOI in \datasetnamenospace, we compare the \textbf{zero-shot} performance of major existing video-language models on MCQ (\cref{subsec:mcq}) and ReasoningVOS (\cref{subsec:rvos}) tasks, and evaluate VLM fine-tuning with additional hand and object cues to find future directions for model development (\cref{subsec:fine-tuning}).


\subsection{Multiple-Choice Questions}
\label{subsec:mcq}

\begin{figure*}[t]
    \centering
    \includegraphics[width=1\linewidth]{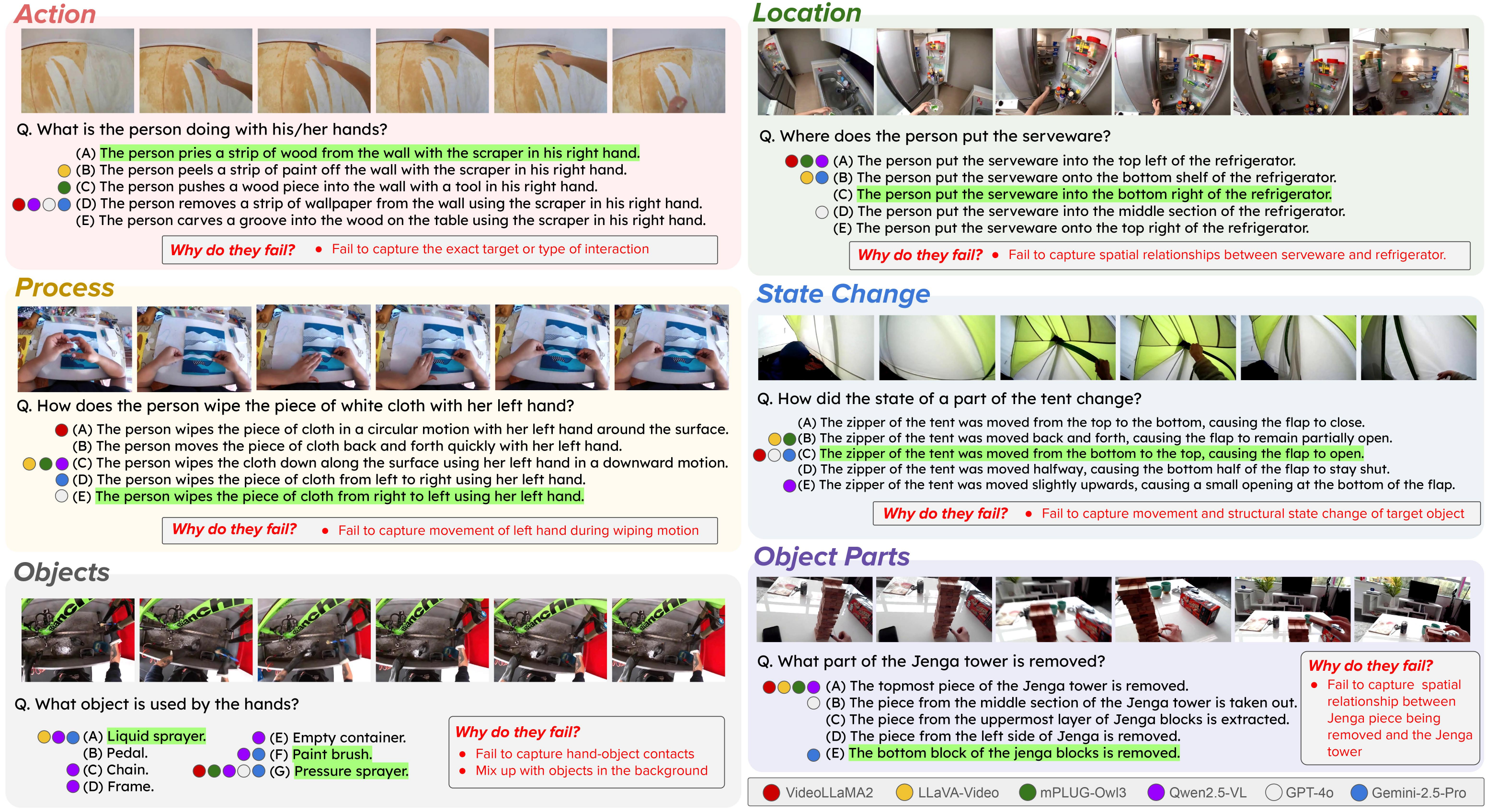}
    \caption{Qualitative results. Sentence with green highlights denote correct answer. \ty{Colored circles denote options predicted by each model.}}
    \label{fig:mcq_qualitative}
\end{figure*}



\begin{figure*}[t]
    \centering
    \begin{subfigure}[t]{0.52\linewidth}
        \centering
        \includegraphics[width=\linewidth]{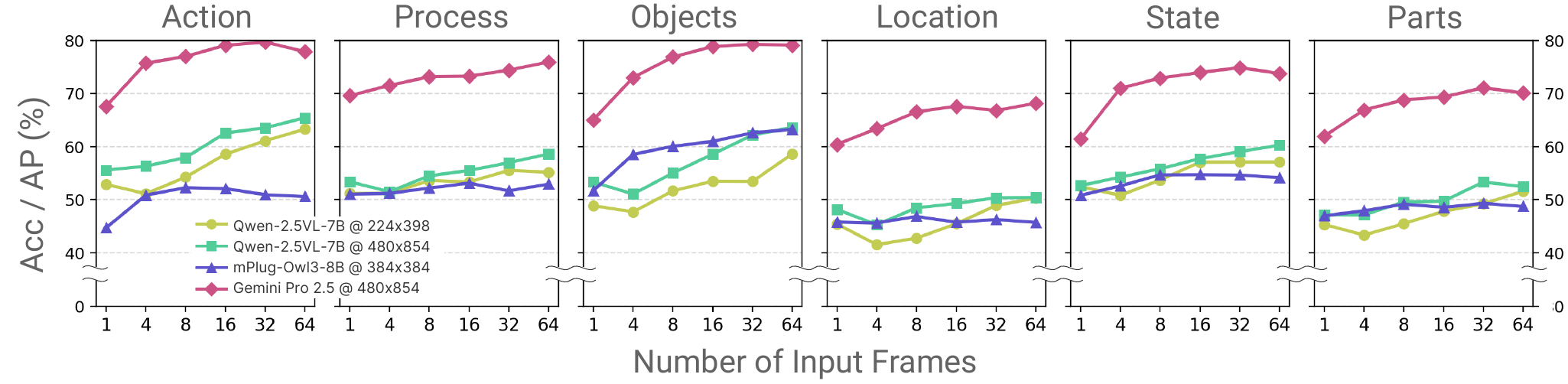}
        \caption{Performance of MCQ by category.}
        \label{fig:frame_number_resolution}
    \end{subfigure}
    \hfill
    \begin{subfigure}[t]{0.46\linewidth}
        \centering
        \includegraphics[width=\linewidth]{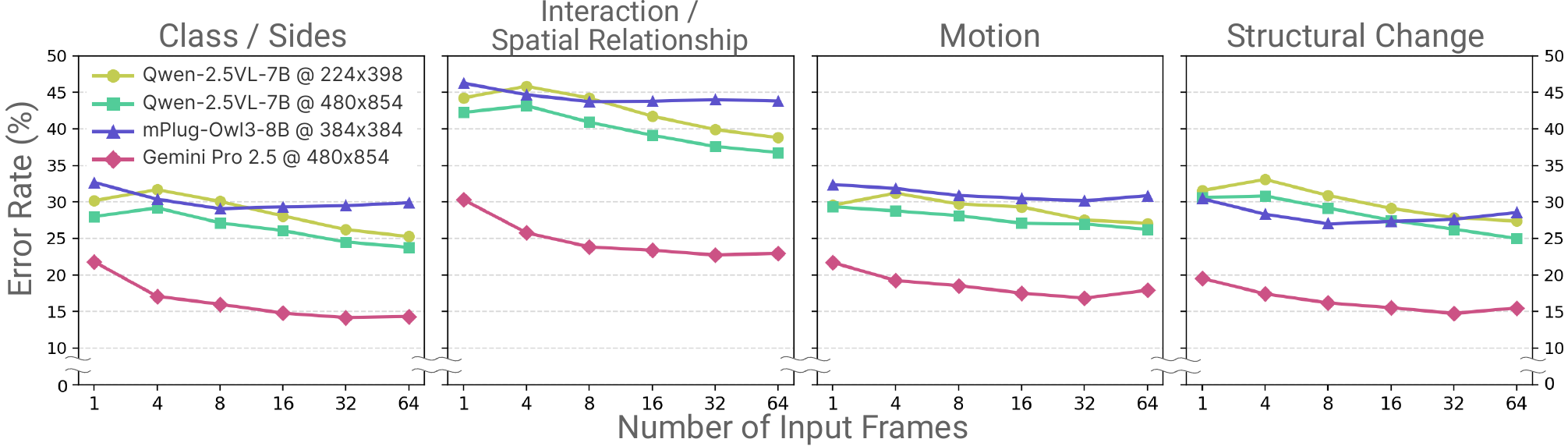}
        \caption{Error rate trends by error type.}
        \label{fig:reason_classification_resolution}
    \end{subfigure}
    \caption{Ablation and error analyses over input frames from 1 ($\approx$0.2~fps) to 64 ($\approx$12.8~fps) and resolutions.}
    \label{fig:frame_reason_combined}
\end{figure*}

\begin{table*}[ht]
    \centering
    \resizebox{\linewidth}{!}{%
    \begin{tabular}{lcc|cccccc|c|cccc}
        \toprule
        \multirow{2}{*}{\textbf{Models}}        & \multirow{3}{*}{\textbf{\#Frames}} & \multirow{3}{*}{\textbf{Key Features}} & \multicolumn{7}{c|}{MCQ Peformance ($\uparrow$)} & \multicolumn{4}{c}{Error Rate ($\downarrow$)} \\
        \cline{4-14} 
        & & &
                               \cellcolor{actColor} \textbf{Action} & \cellcolor{proColor} \textbf{Process} & \cellcolor{objColor} \textbf{Objects}  & \cellcolor{locColor} \textbf{Location} & \cellcolor{staColor} \textbf{State} & \cellcolor{parColor} \textbf{Parts} & \textbf{Avg.} 

                               & \multirow{2}{*}{\textbf{Class/Side}} & \textbf{Int. /} & \multirow{2}{*}{\textbf{Motion}} & \textbf{Structural}
                               \\
                                (Zero-shot) & & & (Acc) & (Acc) & (AP) & (Acc) & (Acc) & (Acc) & (Acc) & & \textbf{Spatial Rel.} & & \textbf{Change}\\
        \midrule\midrule
        LaViLa-L~\cite{zhao2023learning} & 4 & \NA
                               & 59.8 & 39.6 & 67.5 & 35.9 & 38.7 & 35.2 & 41.8 & 33.5 & 51.3 & 35.9 & 38.9 \\
        HelpingHands-L~\cite{zhang2023helping} & 4 & Hand \& Object BBox Inference
                               & 56.9 (\textcolor{\downcolor}{-2.9}) & 36.4 (\textcolor{\downcolor}{-3.2}) & 69.2 (\textcolor{\upcolor}{+1.7}) & 34.5 (\textcolor{\downcolor}{-1.4}) & 39.2 (\textcolor{\upcolor}{+0.5}) & 34.7 (\textcolor{\downcolor}{-0.5}) & 40.3 (\textcolor{\downcolor}{-1.5}) & 35.9 (\textcolor{\downcolor}{+2.4}) & 52.6 (\textcolor{\downcolor}{+1.3}) & 36.6 (\textcolor{\downcolor}{+0.7}) & 39.2 (\textcolor{\downcolor}{+0.3})\\
        \midrule
        LaViLa-L~\cite{zhao2023learning} & 16 & \NA
                               & 61.6 & 40.1 & 68.5 & 36.9 & 38.9 & 35.6 & 42.6 & 32.6 & 50.3 & 35.9 & 38.7\\
        EgoHOD-L~\cite{pei2025modeling} & 16 & Rich Text \& Motion Adapter
                               & 60.2 (\textcolor{\downcolor}{-1.4}) & 37.3 (\textcolor{\downcolor}{-2.8}) & 75.3 (\textcolor{\upcolor}{+6.8}) & 38.0 (\textcolor{\upcolor}{+1.1}) & 42.3 (\textcolor{\upcolor}{+3.4}) & 35.8 (\textcolor{\upcolor}{+0.2}) & 42.7 (\textcolor{\upcolor}{+0.1}) & 33.8 (\textcolor{\downcolor}{+0.3}) & 49.7 (\textcolor{\upcolor}{-1.6}) & 36.4 (\textcolor{\downcolor}{+0.5}) & 36.5 (\textcolor{\upcolor}{-2.4})\\
        \bottomrule
    \end{tabular}%
    }
    \caption{Comparison of zero-shot performance and error rates of models with and without explicit hand/object modeling. Key Features list the components that make the model hand–object aware. Numbers in \textcolor{\upcolor}{Green}/\textcolor{\downcolor}{Red} indicates improvement/degradation over LaViLa.}
    \label{tab:comparison_lavila}
\end{table*}

\paragraph{Baseline models.}
We select six open-source video LLMs and two proprietary models, categorized into (i) dual-encoder models and  (ii) LLM-integrated models.
\textbf{Dual-encoder models} include LaViLa~\cite{zhao2023learning}, a video-language model trained on egocentric videos, and InternVideo2-Stage2~\cite{wang2024internvideo2}, whose visual encoder is pre-trained on large-scale video-text pairs.
\textbf{LLM-integrated models} include VideoLLaMA2.1-7B~\cite{cheng2024videollama}, which specializes in spatio-temporal modeling; LLaVa-Video-7B~\cite{lin2024video}, trained on general and egocentric video datasets; mPLUG-Owl3-8B~\cite{ye2024mplug}, which efficiently processes long image sequences; and Qwen2.5-VL-7B/72B~\cite{wang2024qwen2}, which accepts video inputs with arbitrary resolutions. For proprietary models, we evaluate GPT-4o~\cite{hurst2024gpt}, which is capable of processing image sequences, in both text-only and vision-enabled settings, and Gemini-2.5-Pro~\cite{comanici2025gemini}, a proprietary model capable of processing video files.

\paragraph{Implementation details.}
We uniformly sample 16 frames from each video and use the default input resolution specified for each model. For Gemini-2.5-Pro, we feed the video input at 3.2 fps (16 frames per video).
All models are evaluated in a zero-shot setting.
Since Qwen2.5-VL supports arbitrary resolutions, we align its input with other 7B-scale models for fair comparison.
For the 72B model, however, we use the full resolution to showcase its full capability.
For dual-encoder models,
we compute the cosine similarity between the video feature and the text feature of each option, selecting the one(s) with the highest score.
For the remaining models, we provide the video frames along with a prompt listing all options and infer the most probable option(s).


\paragraph{Evaluation metrics.}
We report top-1 accuracy for all the categories except {\bf Objects}, and Average Precision (AP) for {\bf Objects} because it has more than one answers per question.

\paragraph{Quantitative results.}
\cref{tab:comparison} shows the quantitative results.
Despite preparing answer options unsolvable from text alone, GPT-4o (text) scored slightly above chance (34–51 pts), indicating minor textual bias but insufficient to solve the task.
The dual encoder-based LaViLa trained on Ego4D dataset outperformed InternVideo2-Stage2, particularly in the {\bf Action} and {\bf Objects} categories, surpassing some LLM-integrated models.
However, it showed inferior performance in other categories, suggesting that LaViLa is specialized to recognize actions and objects. 
LLM-integrated models consistently outperformed the text-only baseline, following similar trends observed in general video understanding tasks~\cite{salehi2024actionatlas}.  
Among the 7B-scale models, LLaVA-Video-7B, fine-tuned on Ego4D, achieved the highest average accuracy of 54.8\%, highlighting the benefits of domain-specific adaptation.  
Among open-source models, Qwen2.5-VL-72B, when using high-resolution inputs, achieved the highest overall performance with an average accuracy of 69.9\%.
Gemini-2.5-Pro attained the highest accuracy across all models (72.6\%), substantially surpassing another proprietary model, GPT-4o (vision) (58.9\%).
However, all models showed limited performance with a substantial gap compared to human score (Avg. 96.6\%), suggesting that current video foundation models still have room for improvement in capturing the fine-grained nuances of hand-object interctions.

\paragraph{Qualitative results.}

\cref{fig:mcq_qualitative} shows examples where most models struggled in each category.
Major failure cases include (i) confusing nearby objects or hand sides (ii) failing to capture the spatial relationship between hand-object or object-object pairs (iii) missing motion of objects or hands, (iv) failing to capture the structural or state changes of the hands or objects themselves.
These results suggest that existing models tend to rely on superficial context cues, often missing subtle yet crucial local interactions, movements, and structural changes in the scene.

\paragraph{Effects of input frame rate and resolution.} 
We changed the number of input frames for mPLUG-Owl3 and Gemini-2.5-Pro while keeping the resolution at 384$\times$384 and 480$\times$854, respectively, and varied both the number of input frames and resolutions for Qwen2.5-VL.

As shown in \cref{fig:frame_number_resolution}, increasing both the number of input frames and spatial resolution enhanced performance, with the most consistent gains from increased frame counts observed for \textbf{Gemini-2.5-Pro}. \textbf{Qwen2.5-VL-7B} also benefited from both higher frame counts and resolutions, particularly in \textbf{Action} and \textbf{Objects}. \textbf{Location} benefited from higher resolution only under a small number of input frames, whereas improvements for \textbf{Process} remain limited. In contrast, \textbf{mPLUG-Owl3-8B} saturates beyond 8 frames across all categories except \textbf{Objects}.


\begin{table*}[ht]
    \centering
    \small
    \resizebox{\linewidth}{!}{%
    \begin{tabular}{lll|cccc|cccc|cccc|cccc}
        \toprule
        \multirow{2}{*}{Models} &  \multirow{2}{*}{Input} & \multirow{2}{*}{Prompt}
        & \multicolumn{4}{c|}{\cellcolor{objColor} Objects ($\mathcal{J}$\,$\uparrow$)} 
        & \multicolumn{4}{c|}{\cellcolor{objColor} Objects ($\mathcal{F}$\,$\uparrow$)} 
        & \multicolumn{4}{c|}{\cellcolor{parColor} Parts ($\mathcal{J}$\,$\uparrow$)} 
        & \multicolumn{4}{c}{\cellcolor{parColor} Parts ($\mathcal{F}$\,$\uparrow$)} \\
        & & & S & M & L & All & S & M & L & All & S & M & L & All & S & M & L & All \\
        \midrule
        VideoLISA-3.8B & Video & Question & 4.4 & 10.6 & 7.7 & 7.6 & 5.3 & 10.4 & 8.7 & 8.1 & 0.2 & 1.3 & 4.9 & 2.1 & 0.6 & 1.7 & 3.7 & 2.0 \\

        VideoLISA-3.8B & Video & GT Option & 0.7 & 3.1 & 4.9 & 2.9 & 2.3 & 4.7 & 4.3 & 3.8 & 0.2 & 1.0 & 3.2 & 1.5 & 0.9 & 1.7 & 3.6 & 2.0 \\

        Sa2VA-4B
        & Video & Question & 11.4 & 20.9 & 19.2 & 17.2 & 14.1 & 18.8 & 15.5 & 16.2 & 1.3 & 5.2 & 15.6 & 7.4 & 3.2 & 7.4 & 11.7 & 7.2 \\

        Sa2VA-8B
        & Video & Question & \textbf{22.4} & 38.0 & 35.6 & 32.0 & \textbf{28.0} & \textbf{36.4} & 29.6 & \textbf{31.3} & 1.8 & 7.7 & 23.3 & 10.9 & \textbf{4.1} & \textbf{9.9} & 16.2 & \textbf{10.1} \\
        
        Sa2VA-8B
        & Image & Question & 21.8 & \textbf{42.6} & 44.3 & \textbf{36.2} & 22.8 & 35.1 & \textbf{30.7} & 29.6 & \textbf{1.9} & \textbf{8.5} & \textbf{27.3} & \textbf{12.6} & 3.7 & 9.5 & 16.6 & 9.8 \\
        
        
        Sa2VA-8B
        & Video & GT Option & 17.9 & 35.8 & \textbf{49.7} & 34.5 & 20.8 & 28.8 & 30.1 & 26.7 & 1.5 & 7.1 & 25.8 & 11.5 & 3.2 & 9.0 & \textbf{17.8} & 10.0 \\
        
        \bottomrule
    \end{tabular}
    }
    \caption{Results of ReasoningVOS for \textbf{Objects} and \textbf{Parts} categories. S/M/L denotes groups of videos categorized by the average size of annotated masks within each video. The best score within each mask-size group is highlighted in \textbf{bold}. GT denotes ground-truth.}
    \label{tab:segmentation_combined}
\end{table*}

\paragraph{Error analysis over spatiotemporal resolutions.}
We statistically analyze the underlying causes of recognition errors to better understand the limitations of current video models.
By comparing the incorrect options with the correct ones for each question, we categorize the errors into four types:
(i) \textbf{Class / Side Confusion}, (ii) \textbf{Interaction / Spatial Relationship Error}, (iii) \textbf{Motion Error}, and (iv) \textbf{Structural Change Error}.
Each distractor may correspond to one or more error types.
We annotated the above error types using GPT-4o for all categories except \textbf{Objects}.
The error rate for each model was computed per error type, where the rate for type $X$ is calculated by $\frac{\#\text{questions where the model selected an option of } X}{\#\text{questions containing an option of } X}$.

\cref{fig:reason_classification_resolution} presents the error rate trends across different input frame counts and resolutions for mPLUG-Owl3, Qwen2.5-VL, and Gemini-2.5-Pro.
The \textbf{Interaction / Spatial Relationship} class showed the highest error rate across models, accounting for the lower performance in \textbf{Location} and \textbf{Parts}.
mPLUG-Owl3 achieved its lowest error rate at 8 frames for all categories except \textbf{Motion}, explaining the observed performance saturation. This is likely because the model was trained on eight-frame video clips, whereas Qwen2.5-VL is trained with dynamically varying frame rates and resolutions.
\textbf{Motion} showed the most limited gains from increasing frame counts and resolutions across models, which explains the modest improvement in \textbf{Process} and \textbf{Location}.
In contrast, \textbf{Gemini-2.5-Pro} exhibited the most consistent reduction in error rates across all error classes as the number of input frames increased, achieving the lowest errors at 32 frames.

The above findings suggest that the open source models struggle at (i) capturing spatial relationships between entities over time, and (ii) effectively utilizing increased temporal information during inference.

\paragraph{Does hand/object-aware model help?}
We evaluated HelpingHands~\cite{zhang2023helping} and EgoHOD~\cite{pei2025modeling}, two models explicitly designed to capture hand/object-aware features and trained on egocentric videos. The former extends the LaViLa visual encoder, while the latter builds on CLIP of a similar size. Both models are trained using auxiliary supervision based on hand and object bounding box locations—an approach likely well suited to our HOI benchmark.
The results are shown in \cref{tab:comparison_lavila}.
HelpingHands improves performance in \textbf{Objects} due to bounding box supervision of manipulated objects, but introduces more errors across all error types, leading to a performance drop in other categories.
EgoHOD reduces errors related to spatial relationships and structural changes through richer textual supervision, resulting in improvements in \textbf{Objects}, \textbf{Location}, and \textbf{State}. However, the Motion Adapter failed to effectively mitigate motion-related errors.
These findings suggest that current hand/object-aware modeling has limited impact on distinguishing hand sides, capturing objects in the environment, and understanding motion.

\subsection{Reasoning Video Object Segmentation}
\label{subsec:rvos}

\paragraph{Baseline models.}
We compare two baseline models: VideoLISA~\cite{bai2024one} and Sa2VA-4B/8B~\cite{sa2va}, both of which are multimodal large language models (MLLM) capable of solving referring video object segmentation. Since Sa2VA is also trained for image segmentation, we performed an ablation comparing image and video inputs. The image input processes each frame independently without using temporal context. We evaluated segmentation performance using either the question or the ground-truth option as text prompt.


\paragraph{Evaluation Metrics.}
Following standard VOS evaluation protocols~\cite{pont20172017,xu2018youtube}, we use the Jaccard Index ($\mathcal{J}$) and Boundary F-measure ($\mathcal{F}$) computed for each frame and report their average over annotated frames.
Furthermore, we categorize the videos into three size-based groups (S/M/L) based on the area of the ground truth masks averaged over each video, and report their average scores.

\paragraph{Implementation details.}
We input 16 uniformly sampled frames per video for all models, while ensuring that the annotated frames are included.

\begin{table*}[t]
    \centering
    \resizebox{\linewidth}{!}{%
    \begin{tabular}{lc|cccccc|c|cccc}
        \toprule
        \multirow{3}{*}{\textbf{Input}} & \multirow{3}{*}{\textbf{Fine-tune}} & \multicolumn{7}{c|}{MCQ Peformance ($\uparrow$)} & \multicolumn{4}{c}{Error Rate ($\downarrow$)} \\
        \cline{3-13} 
        & &
                               \cellcolor{actColor} \textbf{Action} & \cellcolor{proColor} \textbf{Process} & \cellcolor{objColor} \textbf{Objects}  & \cellcolor{locColor} \textbf{Location} & \cellcolor{staColor} \textbf{State} & \cellcolor{parColor} \textbf{Parts} & \textbf{Avg.} 

                               & \multirow{2}{*}{\textbf{Class/Side}} & \textbf{Int. /} & \multirow{2}{*}{\textbf{Motion}} & \textbf{Structural}
                               \\
        & & (Acc) & (Acc) & (AP) & (Acc) & (Acc) & (Acc) & (Acc) & & \textbf{Spatial Rel.} & & \textbf{Change}\\
        \midrule
        \midrule
        RGB (Zero-shot)               & No & 41.1 & 30.2 & 37.0 & 29.7 & 34.9 & 30.5 & 33.3 & 42.7 & 59.9 & 42.3 & 44.2\\
        
        \midrule
        RGB                          & Yes & 49.3 & 66.2 & 36.3 & 47.0 & 46.3 & 46.4 & 51.0 & 35.3 & 41.7 & 29.5 & 32.8 \\
        
        RGB + Hand                   & Yes & 53.0 (\textcolor{\upcolor}{+3.7}) & 68.3 (\textcolor{\upcolor}{+2.1}) & 36.2 (\textcolor{\downcolor}{-0.1}) & 48.4 (\textcolor{\upcolor}{+1.4}) & 48.8 (\textcolor{\upcolor}{+2.5}) & 47.9 (\textcolor{\upcolor}{+1.5}) & 53.3 (\textcolor{\upcolor}{+2.3}) & 33.6 (\textcolor{\upcolor}{-1.7}) & 39.5 (\textcolor{\upcolor}{-2.2}) & 27.9 (\textcolor{\upcolor}{-1.6}) & 31.2 (\textcolor{\upcolor}{-1.6})\\
        
        RGB + ObjBBox                    & Yes & 51.0 (\textcolor{\upcolor}{+1.7}) & 68.9 (\textcolor{\upcolor}{+2.7}) & 37.6 (\textcolor{\upcolor}{+1.3}) & 47.0 (\textcolor{black}{+0.0}) & 49.3 (\textcolor{\upcolor}{+3.0}) & 45.9 (\textcolor{\downcolor}{-0.5}) & 52.4 (\textcolor{\upcolor}{+1.4}) & 34.9 (\textcolor{\upcolor}{-0.4}) & 40.4 (\textcolor{\upcolor}{-1.3}) & 28.0 (\textcolor{\upcolor}{-1.5}) & 31.1 (\textcolor{\upcolor}{-1.7})\\
        
        RGB + ObjFeats            & Yes & 51.9 (\textcolor{\upcolor}{+2.6}) & 68.4 (\textcolor{\upcolor}{+2.2}) & \textbf{38.6} (\textcolor{\upcolor}{+2.3}) & 48.2 (\textcolor{\upcolor}{+1.2}) & \textbf{50.3} (\textcolor{\upcolor}{+4.0}) & 47.8 (\textcolor{\upcolor}{+1.4}) & 53.3 (\textcolor{\upcolor}{+2.3}) & 34.1 (\textcolor{\upcolor}{-1.2}) & 39.6 (\textcolor{\upcolor}{-2.1}) & 28.1 (\textcolor{\upcolor}{-1.4}) & 30.4 (\textcolor{\upcolor}{-2.4})\\
        
        RGB + Hand + ObjBBox             & Yes & \textbf{53.8} (\textcolor{\upcolor}{+4.5}) & \textbf{69.6} (\textcolor{\upcolor}{+3.4}) & 37.3 (\textcolor{\upcolor}{+1.0}) & 47.6 (\textcolor{\upcolor}{+0.6}) & 49.3 (\textcolor{\upcolor}{+3.0}) & 47.0 (\textcolor{\upcolor}{+0.6}) & 53.5 (\textcolor{\upcolor}{+2.5}) & 33.8 (\textcolor{\upcolor}{-1.5}) & 39.2 (\textcolor{\upcolor}{-2.5}) & 27.3 (\textcolor{\upcolor}{-2.2}) & 31.3 (\textcolor{\upcolor}{-1.5})\\
        
        RGB + Hand + ObjFeats     & Yes & 52.9 (\textcolor{\upcolor}{+3.6}) & 69.2 (\textcolor{\upcolor}{+3.0}) & 37.8 (\textcolor{\upcolor}{+1.5}) & \textbf{48.7} (\textcolor{\upcolor}{+1.7}) & 49.5 (\textcolor{\upcolor}{+3.2}) & \textbf{49.1} (\textcolor{\upcolor}{+2.7}) & \textbf{53.9} (\textcolor{\upcolor}{+2.9}) & 33.4 (\textcolor{\upcolor}{-1.9}) & 38.8 (\textcolor{\upcolor}{-2.9}) & 27.4 (\textcolor{\upcolor}{-2.1}) & 31.0 (\textcolor{\upcolor}{-1.8})\\
        
        RGB + Hand + ObjBBox + ObjFeats & Yes & 51.2 (\textcolor{\upcolor}{+1.9}) & 69.1 (\textcolor{\upcolor}{+2.9}) & 38.0 (\textcolor{\upcolor}{+1.7}) & 47.9 (\textcolor{\upcolor}{+0.9}) & \textbf{50.3} (\textcolor{\upcolor}{+4.0}) & 48.8 (\textcolor{\upcolor}{+2.4}) & 53.4 (\textcolor{\upcolor}{+2.4}) & 33.9 (\textcolor{\upcolor}{-1.4}) & 39.3 (\textcolor{\upcolor}{-2.4}) & 27.6 (\textcolor{\upcolor}{-1.9}) & 30.8 (\textcolor{\upcolor}{-2.0})\\

        \bottomrule
    \end{tabular}
    }
    \caption{Comparison between different inputs. Numbers in \textcolor{\upcolor}{Green}/\textcolor{\downcolor}{Red} indicates improvement/degradation over RGB input.}
    \label{tab:hoi_cues}
\end{table*}

\paragraph{Results.}
As shown in \cref{tab:segmentation_combined}, all the models performed significantly worse than those in prior ReasoningVOS tasks ({\it e.g.,} 40+ $\mathcal{J}$ with VideoLISA in \cite{bai2024one}). This degradation mainly stems from the increased scene clutter and the fact that the questions require detecting subtle hand–object contacts, unlike in prior Referring or ReasoningVOS benchmarks. In particular, the {\bf Parts} category demands component-level reasoning and segmentation, which is more challenging than object-level segmentation.

\begin{figure}[t]
    \centering
    \includegraphics[width=0.9\linewidth]{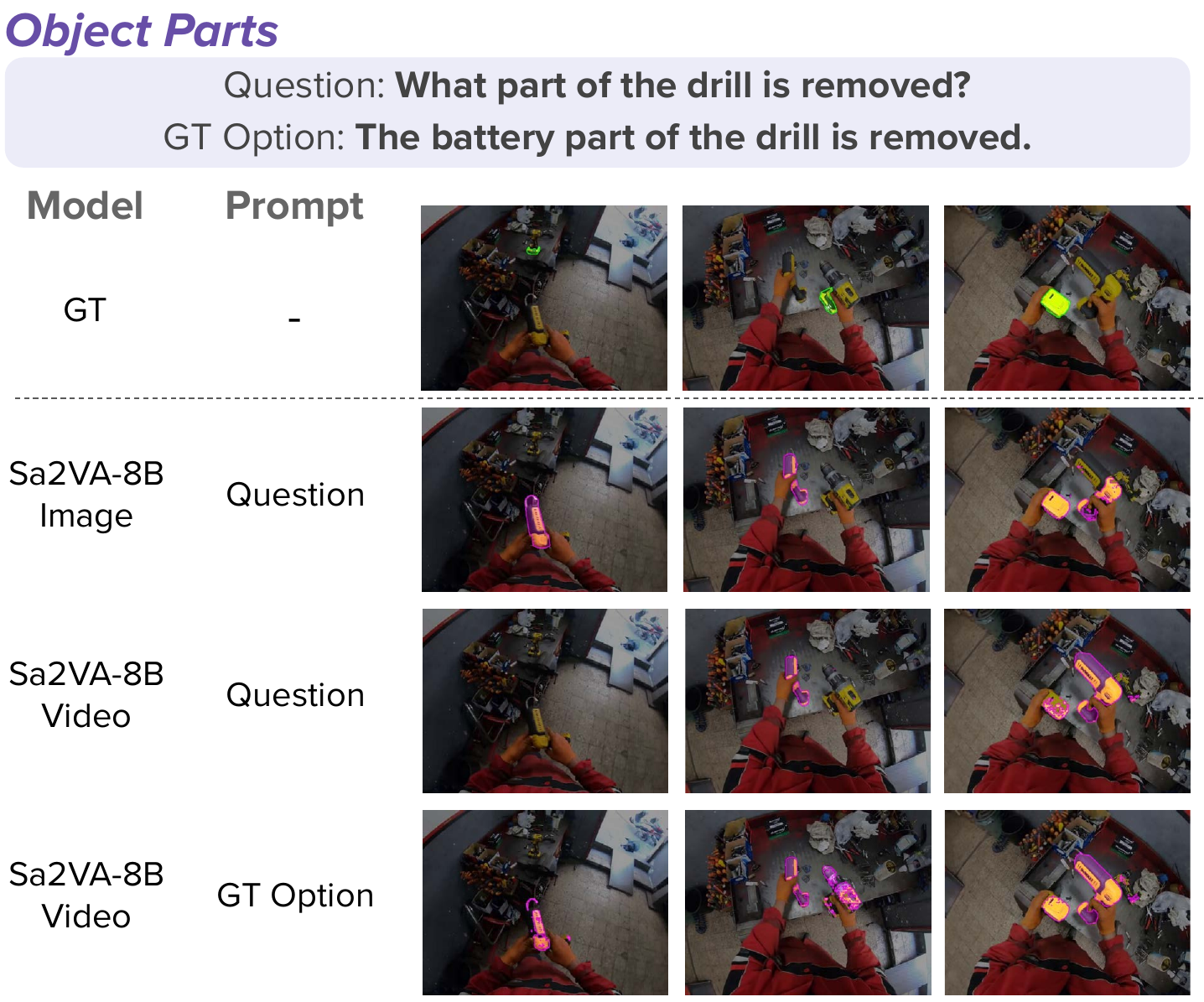}
    \caption{Qualitative results on ReasoningVOS. GT means ground-truth. GT masks are shown in \textcolor{green}{green} in first row, while predictions of each model are shown in \textcolor{magenta}{magenta}. Models mistakenly segment entire object instead of correct part. Note that video models are provided with 16 frames during inference for temporal context.
    }
    \label{fig:seg_qualitative}
\end{figure}

The overall results show that larger models achieve higher scores, indicating the importance of the reasoning capacity of the models.
We observed different trends in each size group.
Providing video input yielded better results than frame-wise processing for larger masks (group L). However, frame-wise processing sometimes performed better for smaller masks, as it can capture small objects without being distracted by temporal context from adjacent frames, as shown in the last frame on the right side of \cref{fig:seg_qualitative}.

Among 8B models, using the GT options as prompts yielded better scores for larger masks, whereas using questions as prompts performed better for smaller masks (groups S and M).
This is possibly because the ground truth text used to prompt MLLMs may be not sufficient to describe the precise region in fine-grained HOIs, often leading to over-segmentation of the target area.

\subsection{Integration of HOI cues}
\label{subsec:fine-tuning}
We also investigate additional factors to better capture the hand-object dynamics in our benchmark.
We hypothesize that explicitly adding spatio-temporally local information about hand manipulations and interacting objects may improve performance. 

To test this hypothesis, we trained different models with additional branches that input combination of additional cues.
We chose InternVideo2-Stage2~\cite{wang2024internvideo2} as our baseline model and fine-tune the model using the training split (1.1~K questions), which could be regarded as a small-scale instruction tuning set.
Specifically, we considered (i) 3D hand pose information, (ii) bounding box tracklets of manipulated objects, and (iii) their object features as additional cues.

\paragraph{Implementation details.}
In addition to the original visual/text encoders, separate encoders for each modality consisting of frame-wise MLP and LSTM are introduced.
These cues are concatenated with the video feature and passed to a projection layer to match the embedding space.
The visual encoder and text encoder of InternVideo2 remain frozen during training, and only the added layers are trained using contrastive loss.
We input 16 frames per video with a resolution of 224$\times$224.
63-dimensional 3D hand poses for both hands are extracted using WiLoR~\cite{potamias2024wilor}.
4-dimensinal bounding box tracklets of manipulated objects are obtained using AMEGO~\cite{goletto2024amego}.
768-dimensional object features are extracted using CLIP~\cite{radford2021learning}. See supplementary for more details.


\paragraph{Results.}
Table~\ref{tab:hoi_cues} shows the comparison across different modalities.
First, we observe significant improvement by applying fine-tuning (51.0 vs. 33.3 avg. accuracy).
The additional cues improved performance across all question types. The error rate results show that 3D hand pose and bounding box tracklets are particularly effective for reducing \textbf{Class / Side} and \textbf{Interaction / Spatial Relationship} errors, while object features mainly mitigate \textbf{Structural Change} errors. \textbf{Motion} errors are alleviated, especially when hand pose is combined with either of the object-related cues.

Performance sometimes improves more with fewer cues, likely due to our simple concatenation design. We leave better modeling and integration strategies for future work.
\section{Conclusion}
We have proposed \datasetnamenospace, a new video QA benchmark for evaluating spatiotemporal dynamics in HOIs.
Experimental results reveal that strong video-language models struggle in capturing the fine-grained dynamics of HOIs, only achieving at most 68--79\% top-1 accuracy in MCQ, and showing poor performance in grounding local regions.
Ablation studies and modality analysis show that increasing the number of input frames and resolutions, as well as incorporating explicit HOI cues, improves overall performance.
These findings suggest that future models should more effectively exploit richer spatio-temporal inputs and explicitly model local hand-object dynamics, rather than relying on a fixed ViT backbone pretrained on low-resolution videos (e.g., 8 fps at 224x224).
We hope that \datasetname~opens up new avenues for advancing deeper HOI understanding.

\noindent\textbf{Acknowledgement.}
This work was supported by JST ASPIRE Grant Number JPMJAP2303, JST K Program Grant Number JPMJKP25V1, and the AIST policy-based budget project ``R\&D on Generative AI Foundation Models for the Physical Domain''. We used ABCI 3.0 provided by AIST and AIST Solutions with support from ``ABCI 3.0 Development Acceleration Use''.
We thank BAOBAB Inc. for their contribution to providing human evaluations for our dataset.
We also thank Wataru Kawabe and Yuki Sakai for their help with annotation checking.

{
    \small
    \bibliographystyle{ieeenat_fullname}
    \bibliography{main}
}


\clearpage
\maketitlesupplementary
\appendix
\setcounter{page}{1}
\renewcommand{\contentsname}{Supplementary Material Contents}

\section*{Supplementary Material Contents}
\begin{enumerate}
  \renewcommand{\labelenumi}{\Alph{enumi}.}
  \item Details on Experimental Results \dotfill \pageref{sec:details_on_results}
  \item Details on Data Collection \dotfill \pageref{sec:details_on_data_collection}
  \item Details on Experimental Settings \dotfill \pageref{sec:details_on_experimental_settings}
  \item Fine-tuning Qwen2.5-VL-7B \dotfill \pageref{sec:fine-tune_qwen}
  \item Broader Impacts \dotfill \pageref{sec:broader_impacts}
  
\end{enumerate}


\section{Details on Experimental Results}
\label{sec:details_on_results}
\subsection{Multiple-Choice Questions}
\paragraph{Performance comparison by GPT-4o answerability.}
In multiple-choice questions, the GPT-4o (text/vision) models often declined to answer, seemingly when the model evaluated that there is insufficient information to respond to the question, even though we asked it to answer the question.
Reporting results only for the questions that GPT-4o answered risks an unfair comparison, as those questions might also be easy for the other models.
To this end, we re-evaluated every model on the subset of questions answered by both the text and vision variants of GPT-4o, reducing the valid items from 9,431 to 6,629 (70.3\%).

Table \ref{tab:comparison_with_deltas} reports these subsampled scores together with their deltas from the original evaluation.
Most metrics increase, indicating that GPT-4o tends to skip more difficult questions that other models also frequently fail on.
In contrast, human performance remains largely unchanged across categories (within 0.5 points), suggesting that humans can reliably recognize HOI dynamics even in challenging videos where MLLMs struggle.

We divided the questions into four categories according to whether the GPT-4o text and vision variants produced an answer, then evaluated Qwen2.5-VL-72B on each category (Table \ref{tab:comparison_if_answer}). Accuracy peaked for questions answered by both GPT-4o variants and declined whenever either variant abstained. This pattern suggests that GPT-4o can recognize when textual or visual information is insufficient and refrain from answering, thereby avoiding errors.
This also suggests that models specifically designed for videos perform better than the general-purpose GPT-4o model, showing that our benchmark poses a challenging video QA task.

\begin{table*}[t]
    \raggedright
    \resizebox{\linewidth}{!}{%
    \begin{tabular}{lccc|cccccc|c}
        \toprule
        \textbf{Models} & \multirow{2}{*}{\textbf{Visual Backbone}} & \multirow{2}{*}{\textbf{Resolution}} & \multirow{2}{*}{\textbf{LLM}} & \cellcolor{actColor} \textbf{Action} & \cellcolor{proColor} \textbf{Process} & \cellcolor{objColor} \textbf{Objects}  & \cellcolor{locColor} \textbf{Location} & \cellcolor{staColor} \textbf{State} & \cellcolor{parColor} \textbf{Parts} & \textbf{Avg.} \\
         (Zero-shot) & & & & (Acc) & (Acc) & (AP) & (Acc) & (Acc) & (Acc) & (Acc)  \\
        \midrule
        \midrule

        \multicolumn{3}{l}{\textit{\textcolor{gray}{Text only models}}} &&&&&&&&\\
        GPT-4o (text)~\cite{hurst2024gpt} & \NA & \NA & GPT-4o  & 37.3 (\textcolor{\upcolor}{+0.7}) & 54.2 (\textcolor{\upcolor}{+3.3}) & 34.4 (\textcolor{\upcolor}{+0.1}) & 36.5 (\textcolor{\upcolor}{+2.4}) & 40.0 (\textcolor{\upcolor}{+0.5}) & 47.5 (\textcolor{\upcolor}{+2.0}) & 43.1 (\textcolor{\upcolor}{+1.8}) \\
        \midrule
        \multicolumn{3}{l}{\textit{\textcolor{gray}{Open-source dual-encoder video-language models}}} &&&&&&&&\\
        LaViLa (TSF-L)~\cite{zhao2023learning} & TimeSformer & 224x224 & \NA & 63.7 (\textcolor{\upcolor}{+2.1}) & 41.7 (\textcolor{\upcolor}{+1.6}) & 68.6 (\textcolor{\upcolor}{+0.1}) & 41.5 (\textcolor{\upcolor}{+4.6}) & 40.6 (\textcolor{\upcolor}{+1.7}) & 37.9 (\textcolor{\upcolor}{+2.3}) & 45.1 (\textcolor{\upcolor}{+2.5}) \\

        InternVideo2-Stage2~\cite{wang2024internvideo2} & Original & 224x224 & \NA  & 42.1 (\textcolor{\upcolor}{+1.0}) & 30.7 (\textcolor{\upcolor}{+0.5}) & 37.1 (\textcolor{\upcolor}{+0.1}) & 34.0 (\textcolor{\upcolor}{+4.3}) & 36.2 (\textcolor{\upcolor}{+1.3}) & 31.3 (\textcolor{\upcolor}{+0.8}) & 34.8 (\textcolor{\upcolor}{+1.5}) \\

        \multicolumn{3}{l}{\textit{\textcolor{gray}{Open source video-language models w/ integrated LLMs}}} &&&&&&&&\\
        VideoLLaMA2.1-7B~\cite{cheng2024videollama} & SigLIP & 384x384 & Qwen2 & 44.0 (\textcolor{\upcolor}{+2.6}) & 50.3 (\textcolor{\upcolor}{+3.0}) & 53.2 (\textcolor{\upcolor}{+0.3}) & 39.3 (\textcolor{\upcolor}{+4.7}) & 47.9 (\textcolor{\upcolor}{+0.9}) & 44.8 (\textcolor{\upcolor}{+4.3}) & 45.2 (\textcolor{\upcolor}{+3.1}) \\
        
        LLaVa-Video-7B~\cite{lin2024video} & SigLIP & 384x384 & LLaVa-7B & 61.3 (\textcolor{\upcolor}{+4.4}) & 58.2 (\textcolor{\upcolor}{+4.5}) & 60.7 (\textcolor{\upcolor}{+0.3}) & 55.4 (\textcolor{\upcolor}{+4.9}) & 60.1 (\textcolor{\upcolor}{+1.6}) & 59.6 (\textcolor{\upcolor}{+5.0}) & 58.9 (\textcolor{\upcolor}{+4.1}) \\
        
        mPLUG-Owl3-8B~\cite{ye2024mplug} & SigLIP & 384x384 & Qwen2 & 55.2 (\textcolor{\upcolor}{+3.1}) & 56.1 (\textcolor{\upcolor}{+3.0}) & 61.1 (\textcolor{\upcolor}{+0.1}) & 50.2 (\textcolor{\upcolor}{+4.4}) & 57.5 (\textcolor{\upcolor}{+2.8}) & 52.6 (\textcolor{\upcolor}{+4.0}) & 54.3 (\textcolor{\upcolor}{+3.4}) \\
        

        Qwen2.5-VL-7B~\cite{wang2024qwen2} & Original & 384x384 & Qwen2.5 & 64.3 (\textcolor{\upcolor}{+3.5}) & 58.5 (\textcolor{\upcolor}{+3.6}) & 54.1 (\textcolor{\upcolor}{+0.2}) & 53.7 (\textcolor{\upcolor}{+5.8}) & 57.9 (\textcolor{\upcolor}{+1.2}) & 53.1 (\textcolor{\upcolor}{+4.5}) & 57.5 (\textcolor{\upcolor}{+3.7}) \\


        Qwen2.5-VL-72B~\cite{wang2024qwen2} & Original & 480x854 & Qwen2.5 & 80.5 (\textcolor{\upcolor}{+2.5}) & 77.6 (\textcolor{\upcolor}{+4.2}) & 75.4 (\textcolor{\upcolor}{+0.2}) & 69.9 (\textcolor{\upcolor}{+6.7}) & 74.4 (\textcolor{\upcolor}{+2.2}) & 66.5 (\textcolor{\upcolor}{+4.0}) & 73.8 (\textcolor{\upcolor}{+3.9}) \\
        \midrule
        \multicolumn{3}{l}{\textit{\textcolor{gray}{Proprietary vision and language models w/ integrated LLMs}}} &&&&&&&&\\
        GPT-4o (vision)~\cite{hurst2024gpt} & Original & 480x854 & GPT-4o & 60.5 (\textcolor{\downcolor}{-0.8}) & 64.6 (\textcolor{\upcolor}{+0.2}) & 64.1 (0.0) & 53.9 (\textcolor{\upcolor}{+2.4}) & 59.4 (\textcolor{\upcolor}{+0.4}) & 59.3 (\textcolor{\upcolor}{+0.8}) & 59.6 (\textcolor{\upcolor}{+0.7}) \\

        Gemini-2.5-Pro~\cite{comanici2025gemini} & Original & 480x854 & Gemini-2.5-Pro & 81.3 (\textcolor{\upcolor}{+2.2}) & 77.9 (\textcolor{\upcolor}{+4.6}) & 78.9 (\textcolor{\upcolor}{+0.1}) & 73.2 (\textcolor{\upcolor}{+5.6}) & 75.1 (\textcolor{\upcolor}{+1.2}) & 74.7 (\textcolor{\upcolor}{+5.4}) & 76.4 (\textcolor{\upcolor}{+3.8}) \\
        \midrule
        Human & \NA & \NA & \NA 
        & 98.6 (\textcolor{\downcolor}{-0.1})
        & 95.9 (\textcolor{black}{0.0})
        & 96.0 (\textcolor{\upcolor}{+0.1})
        & 96.6 (\textcolor{\downcolor}{-0.2})
        & 95.3 (\textcolor{\downcolor}{-0.2})
        & 96.9 (\textcolor{\downcolor}{-0.5})
        & 96.6 (\textcolor{\downcolor}{-0.3}) \\
        \bottomrule
    \end{tabular}
    }
    \caption{Comparison of different models on subset of questions answered by both the text and vision versions of GPT-4o. Only 6.6K questions (70.3\%) are used for this evaluation. Differences from the full results in Table~\ref{tab:comparison} are indicated in \textcolor{\upcolor}{+X.X}/\textcolor{\downcolor}{-X.X}.}
    \label{tab:comparison_with_deltas}
\end{table*}

\begin{table*}[t]
\begin{tabular}{p{1.4cm}p{1.4cm}|cccccc|c}
\toprule
\multicolumn{2}{l|}{Answered by GPT-4o?} & \multicolumn{7}{c}{Results of Qwen2.5-VL-72B (with number of questions)}\\
\textbf{Text} & \textbf{Vision} & \cellcolor{actColor} \textbf{Action} & \cellcolor{proColor} \textbf{Process} & \cellcolor{objColor} \textbf{Objects}  & \cellcolor{locColor} \textbf{Location} & \cellcolor{staColor} \textbf{State} & \cellcolor{parColor} \textbf{Parts} & \textbf{Avg. (Sum)} \\
\midrule
\textcolor{\upcolor}{Yes}  & \textcolor{\upcolor}{Yes}  & 80.5 (955) & 77.6 (1042) & 75.4 (1493) & 69.9 (764) & 74.4 (1319) & 66.5 (1056) & 73.8 (5136) \\
\textcolor{\upcolor}{Yes}  & \textcolor{\downcolor}{No} & 67.6 (136) & 65.5 (521) & 54.5 (12) & 54.8 (361) & 62.6 (211) & 55.5 (373) & 61.2 (1602) \\
\textcolor{\downcolor}{No} & \textcolor{\upcolor}{Yes}  & 78.2 (444) & 50.0 (8) & \NA~ (0) & 63.4 (292) & 54.3 (35) & 50.8 (61) & 59.3 (840) \\

\textcolor{\downcolor}{No} & \textcolor{\downcolor}{No} & 67.6 (105) & 71.4 (7) & \NA~ (0) & 50.6 (170) & 53.8 (13) & 47.2 (53) & 58.1 (348) \\
\bottomrule
\end{tabular}
\caption{Comparison of performance on questions grouped by whether the GPT-4o text/vision models provided an answer. The number in parentheses indicates the number of questions.}
\label{tab:comparison_if_answer}
\end{table*}

\begin{figure*}[t]
    \centering
    \includegraphics[width=1\linewidth]{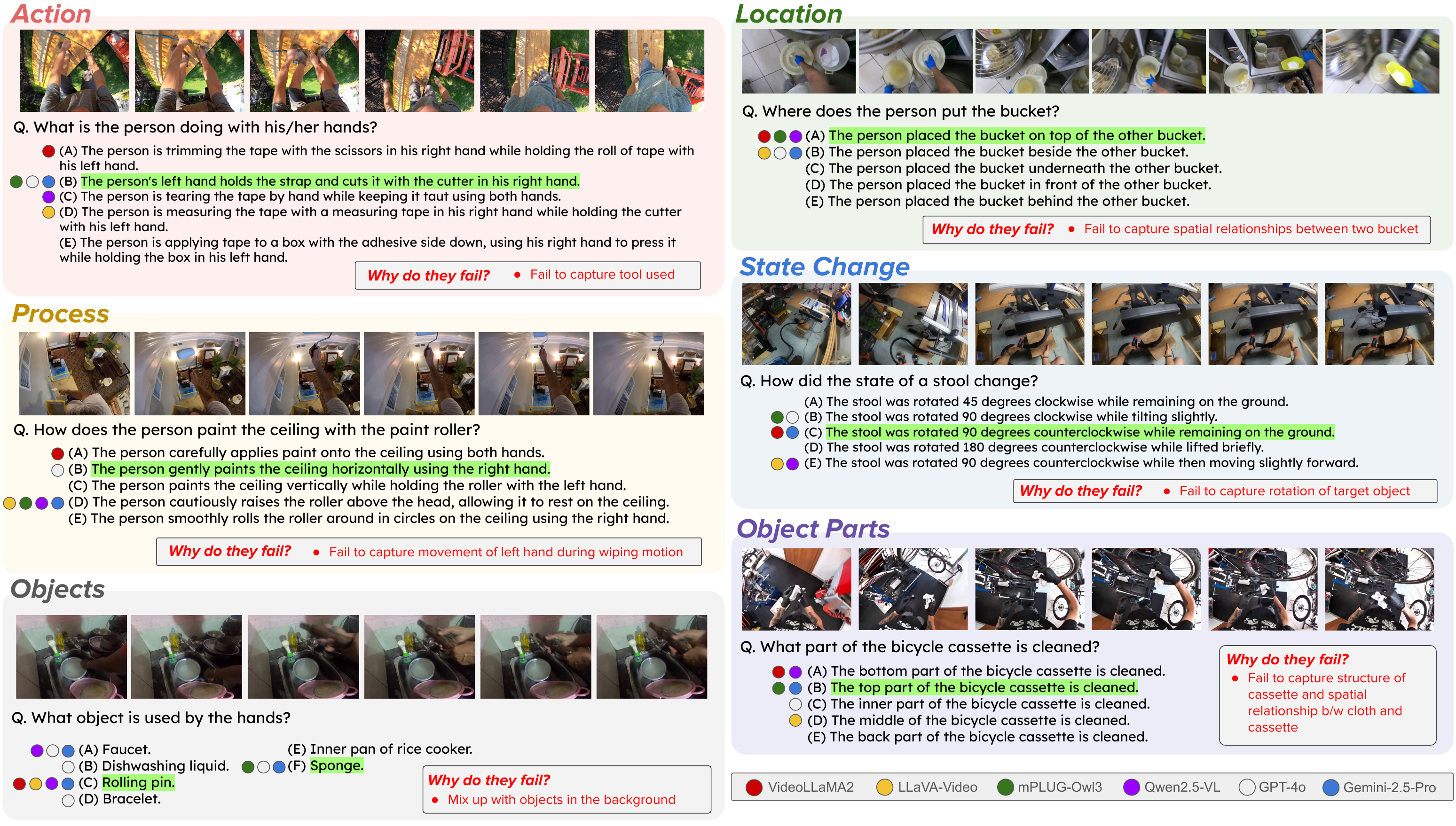}
    \caption{Additional qualitative results for multiple-choice questions. Green highlights denote correct answers.}
    \label{fig:mcq_qualitative_2}
\end{figure*}

\paragraph{Additional Qualitative Results.}
Additional qualitative results per category are shown in Figure~\ref{fig:mcq_qualitative_2}. We observe that current models struggle to accurately recognize objects being manipulated, their spatial relationships with hands or other objects, and their movements.
The original video clips can be found in the supplementary materials.

\begin{figure}[t]
    \centering
    \includegraphics[width=1\linewidth]{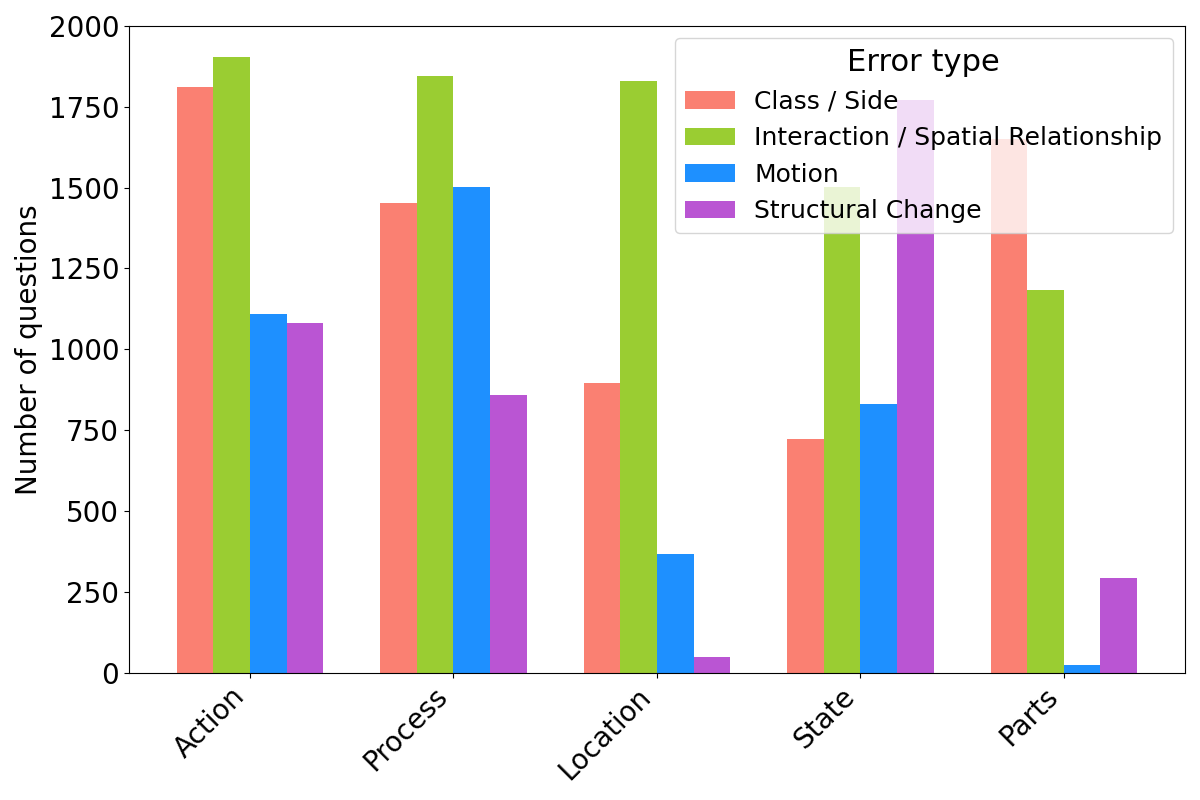}
    \caption{Distribution of error types by question category. Bars indicate the number of questions that contain at least one distractor of each error type.}
    \label{fig:distractor_analysis}
\end{figure}

\paragraph{Error Type Statistics.}
For each distractor, we annotated the following error types by comparing it with the correct answer:

\begin{itemize}
\item \textbf{Class / Side Confusion}: The model confuses the hand side (left or right) or the object class.
\item \textbf{Interaction / Spatial Relationship Error}: The model fails to capture interaction types and spatial relations in hand-object and object-object pairs.
\item \textbf{Motion Error}: The model fails to capture motion or direction over time.
\item \textbf{Structural Change Error}: The model fails to perceive state or structural change in hands/objects.
\end{itemize}

Multiple error types may apply to a single distractor. For example, if the correct answer is ``\textit{He is hammering a cylinder with the hammer in his right hand in a downward motion,}'' and a distractor is ``\textit{He is hammering a cylinder with the hammer in his left hand in a sideways motion,}’’ the assigned error types would be \textbf{Class / Side} and \textbf{Motion}.

Figure~\ref{fig:distractor_analysis} shows the number of questions that include each error type across all question categories. \textbf{Class / Side} and \textbf{Interaction / Spatial Relationship} errors are widely distributed across categories. \textbf{Motion} errors frequently appear in \textbf{Action}, \textbf{Process}, and \textbf{State}, where hand and object movements play critical roles. \textbf{Structural Change} errors appear most often in \textbf{State}, followed by \textbf{Action} and \textbf{Process}, where hand poses are particularly informative. \textbf{Location} contains fewer \textbf{Motion} errors, as distractors primarily describe positional changes using spatial relations between objects. \textbf{Parts} contains fewer \textbf{Structural Change} errors because the LLMs used during annotation often interpret a component as a single object, labeling part-confusion errors as \textbf{Class / Side} confusion instead.

\subsection{Referring Video Object Segmentation}
\begin{figure*}[ht]
    \centering

    \includegraphics[width=\linewidth]{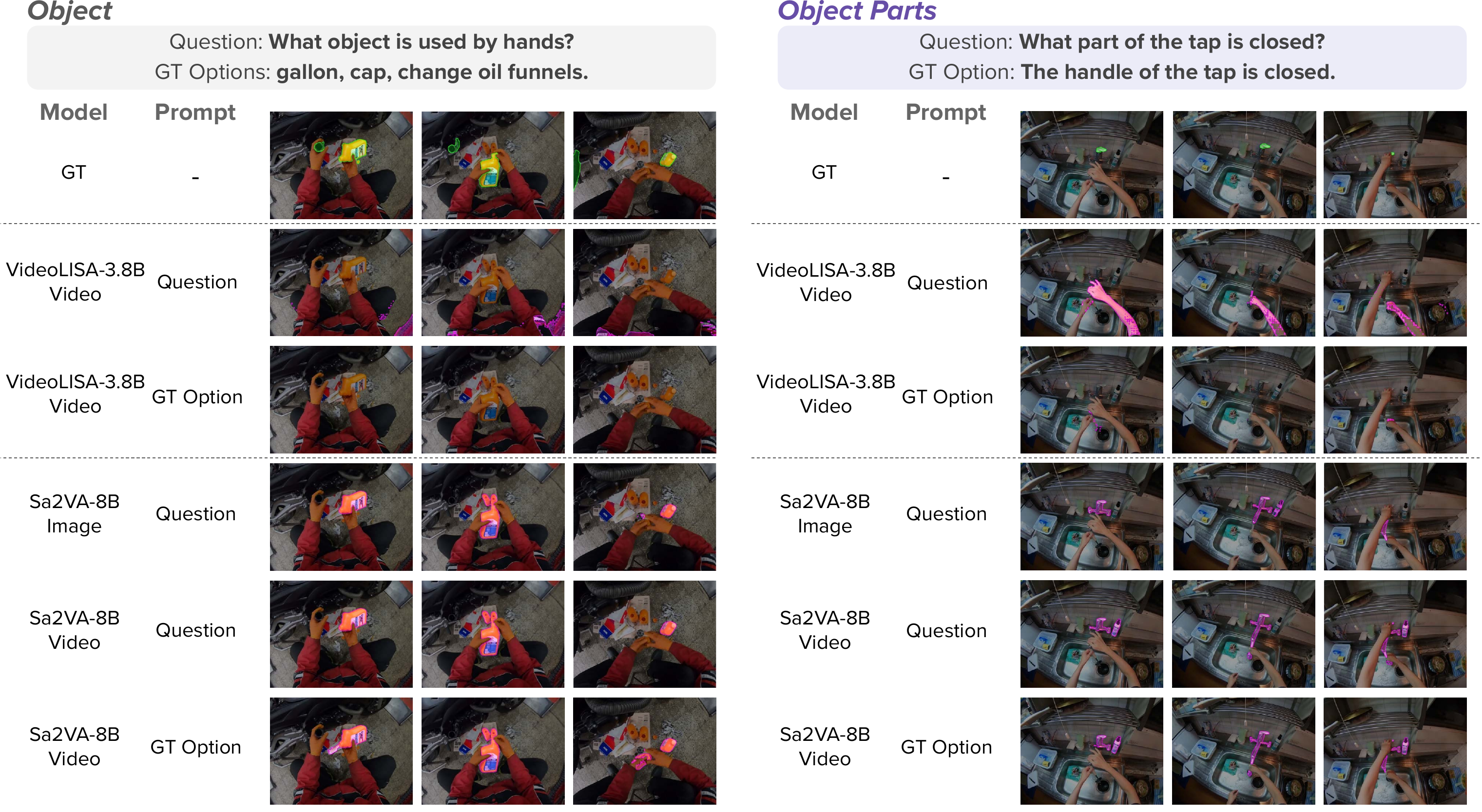}

    \vspace{1em}

    \includegraphics[width=\linewidth]{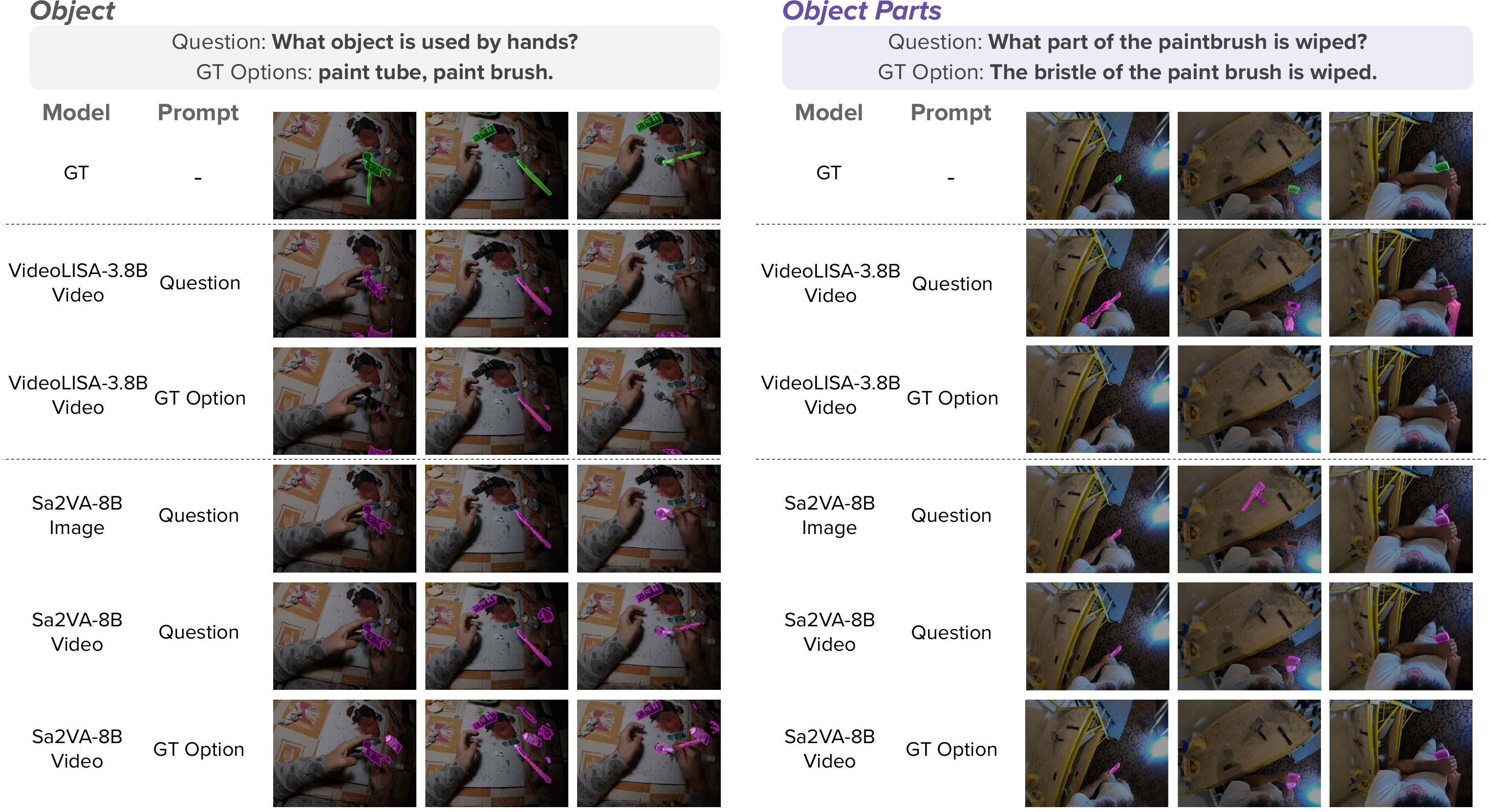}

    \caption{
        Qualitative results on ReasoningVOS. GT means ground-truth. GT masks are shown in \textcolor{green}{green} in first row, while predictions of each model are shown in \textcolor{magenta}{magenta}. 
    }
    \label{fig:seg_qualitative_combined}
\end{figure*}

\paragraph{Additional qualitative results.}
Additional qualitative examples are shown in Figure~\ref{fig:seg_qualitative_combined}.
VideoLISA often fails to detect the target objects even when the ground-truth options are provided as prompts, likely due to domain shift from its training data to egocentric video~\cite{liang2025fine}. It also tends to segment hands instead of the intended object regions in \textbf{Parts} questions.

Sa2VA generally follows the prompts more faithfully in both the \textbf{Objects} and \textbf{Parts} categories. However, when applied in a frame-wise manner, it sometimes segments visually similar but unmanipulated objects, or loses track of the manipulated object because it lacks temporal context (\eg,  missing the paint tube in the bottom-left case, or producing a false-positive segmentation of the paintbrush in the bottom-right example of Figure~\ref{fig:seg_qualitative_combined}). When the ground-truth option is provided, Sa2VA often segments objects of the same class that are not being manipulated. This is likely because the ground-truth description does not precisely specify which instance is being interacted with. The original video clips can be found in the supplementary materials.

\clearpage
\section{Details on Data Collection}
\label{sec:details_on_data_collection}





\paragraph{Details on video clip filtering.}
First, all the narration from stereo videos is filtered out. If a narration is duplicated, it is counted as a single entry. A narration is also filtered out if it does not mention either ``right hand'' or ``left hand.'' Additionally, if the narration contains the word ``unsure'', indicating an unknown object name, it is filtered out. 
We input each narration into large language models (LLMs) to infer the contact objects and secondary objects for each hand (The prompt is shown in Table~\ref {tab:prompt_hoi}). If we can confirm that the camera wearer is manipulating at least one object, we retain the narration for use.
However, if the extracted contact objects include any of the wearer’s body parts or the camera itself, the narration is discarded, as such cases are likely to obscure the visibility of hand-object interactions.

\begin{table}[h]
    \centering
    \begin{tcolorbox}
    
    \small{\textbf{System Prompt}:\\
    You are a helpful assistant who understands interactions between human hands and objects.\\
    \\
    \textbf{User Prompt}:\\
    Please analyze the narration and answer the contact object and the secondary object of each right/left hand. "contact object" means the object that a hand is contacting (manipulating). "secondary object" means the object that the contact object is contacting (the object that is manipulated by the contact object). If the information is not specified in the narration, fill with None.
The answer format should be json:\\
\{“Contact object of right hand”: <obj/None>, “Secondary object of right hand”: <obj/None>, “Contact object of left hand”: <obj/None>, “Secondary object of left hand”: <obj/None>\}\\
\\
Narration: \textcolor{blue}{\{narration text\}}
    }
    \end{tcolorbox}
    \caption{Prompt to extract hand-object interaction information from narration. We replace \textcolor{blue}{\{narration text\}} with narration from Ego4D.}
    \label{tab:prompt_hoi}
\end{table}

\paragraph{Sampling HOI narrations.}
To ensure diverse narration samples for each question, we first sort the remaining narrations chronologically. Then, for each question, we offset the starting index and select every sixth narration. This staggered sampling prevents identical questions from being generated for temporally adjacent HOI segments.

Next, we filter out unsuitable narrations specifically for \textbf{Location} and \textbf{Object Parts}. For \textbf{Location} questions, we select videos where the narration contains verbs indicating object movement. For \textbf{Object Parts} questions, we determine whether the object is partially affected and retain only those where partial impact is evident. This filtering process is done automatically using LLMs based on the narration (The prompt is shown in Table~\ref {tab:prompt_filter_parts}).

Finally, to ensure diversity in HOI samples, we extract verbs from the narrations and randomly select 2,000 samples while ensuring that no single verb appears more than $N$ times. $N$ is determined for each question type to maintain diversity while ensuring at least 2,000 candidate samples are available.

\begin{table}[h]
    \centering
    \begin{tcolorbox}
    
    \small{\textbf{System Prompt}:\\
    You are a helpful assistant who understands interactions between human hands and objects.\\
    \\
    \textbf{User Prompt}:\\
    Create a question for VQA about Hand-Object Interaction.
Specifically, the question should follow the format "What part of [Object] is [Verb]?"\\
(e.g., "What part of the bicycle is replaced?").\\
\\
Given a narration describing a Hand-Object Interaction:\\
1. First, determine if it's an appropriate scene to create a question.\\
2. An appropriate scene should include the following conditions:\\
    - The person is interacting with the object(s).\\
    - The action only affects a limited area of the object 
        (e.g., replacing the tire of a bicycle or folding the left top corner of the paper) 
        rather than the entire object (e.g., moving the speaker).\\
\\
If the scene is appropriate for creating the question, create the question accordingly. If it's inappropriate to create a question, reply with None.\\
If possible, create the corresponding answer to the question. If it's difficult to create the answer, reply with Ambiguous.\\
\\
Here is the narration:\\
\textcolor{blue}{\{narration text\}}\\
\\
Write reasoning and then output the json answer like this:
<reasoning about whether it is appropriate to create a question>\\

```json\\
\{"question": <created question or None>, "answer": <created answer or Ambiguous>\}\\
```\\
    }
    \end{tcolorbox}
    \caption{Prompt to extract hand-object interaction information from narration. We replace \textcolor{blue}{\{narration text\}} with narration from Ego4D.}
    \label{tab:prompt_filter_parts}
\end{table}

\paragraph{Human QA annotation.}
For the automatically generated questions, annotators perform the following tasks while referencing the video:  
(i) verifying the validity of the question,  
(ii) creating the correct answer, and  
(iii) generating incorrect answer choices. 

If an automatically generated question does not match the video, annotators either revise the question or reject the sample. Next, they create the correct answer, ensuring that it provides sufficient detail for the question to be understandable without watching the video.

Once the question and correct answer are prepared, an initial set of plausible incorrect answer choices is generated using LLMs (The prompts are shown in Table~\ref {tab:prompt_option_generation_action}--~\ref {tab:prompt_option_generation_part}). Annotators then review these choices, filtering out inappropriate ones, such as those that overlap with the correct answer. They refine and add challenging distractors that effectively assess comprehension of the question.  

\begin{figure*}[ht]
    \centering
    \includegraphics[width=1\linewidth]{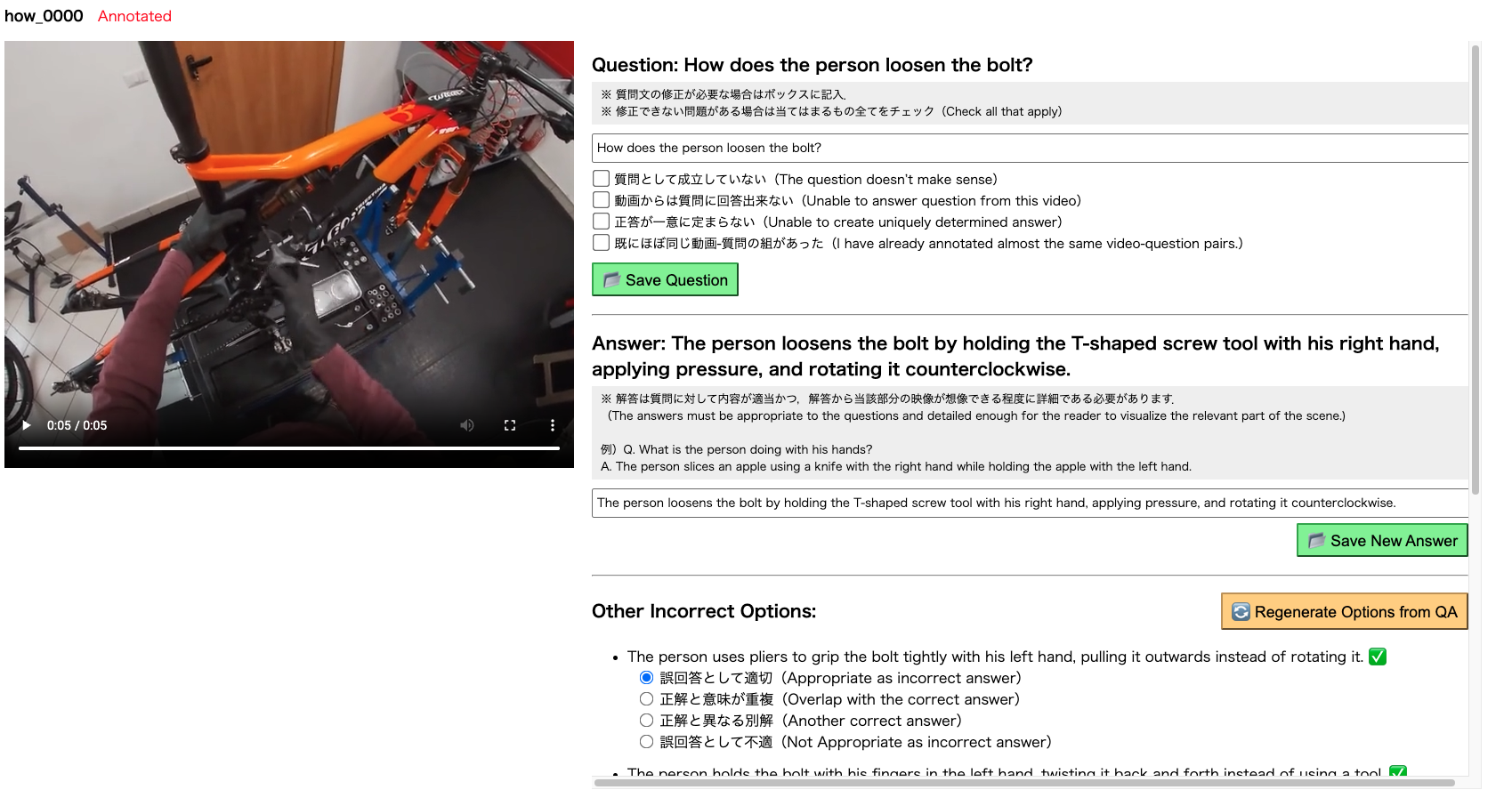}
    \caption{Screenshot of the annotation tool interface. Annotators proceed from top to bottom, sequentially annotating question, correct answer, and distractor options.}
    \label{fig:annot_tool}
\end{figure*}

Overall, human annotators verify all questions, correct answers, and incorrect answer choices, ensuring that the dataset remains sufficiently challenging while still solvable by humans.
The instructions provided to the annotators are shown in Table~\ref{tab:instruction_action}--~\ref{tab:instruction_parts} for each category. The screenshot of the annotation tool interface is provided in Figure~\ref{fig:annot_tool}.

\begin{table}[h]
    \centering
    \begin{tcolorbox}
    \small{
    \textbf{User Prompt}:\\
    You will be given a pair of a question and an answer about a hand-object interaction.\\
Create 4 sentences that describe similar but incorrect answers, as per the examples.\\
Examples:\\
- Question: What is the person doing with his/her hands?\\
- Correct Answer: The person is cutting an apple on the chopping board using the knife in his right hand while holding the apple with his left hand.\\
- Incorrect Answer (object name or situation is wrong): The person is slicing an apple on the table using an apple slicer in his right hand while holding the apple with his left hand.\\
- Incorrect Answer (action is different): The person is removing the skin of an apple on the chopping board using the knife in his right hand while grasping the apple with his left hand.\\

The sentences should:\\
- Contain either different object names or situations (as in the first incorrect example)\\
- Or, contain the same object names but describe different actions (as in the second incorrect example)\\
- Maintain the same level of detail and be of a similar length to the correct answer\\
- Avoid overlapping with each other\\

The answer format should be json:\\
\{"1": "<sentence1>", "2": "<sentence2>", "3": "<sentence3>", ..., "4": "<sentence4>"\}\\

Make sure that you only answer json.\\
\\
Question: \textcolor{blue}{\{question\}}\\
Correct Answer: \textcolor{blue}{\{answer\}}
}
    \end{tcolorbox}
    \caption{Prompt to generate options for \textbf{Action} question. We replace \textcolor{blue}{\{question\}} and \textcolor{blue}{\{answer\}} with created question and answer, respectively}
    \label{tab:prompt_option_generation_action}
\end{table}

\begin{table}[h]
    \centering
    \begin{tcolorbox}
    \small{
    \textbf{User Prompt}:\\
    You will be given a pair of a question and an answer about a hand-object interaction.\\
Create 4 sentences that describe similar but incorrect answers, as per the examples.\\
Examples:\\
- Question: How does the person loosen the fastener?\\
- Correct Answer: The person loosens the fastener by holding the piece firmly in one hand while using the wrench in the other hand to turn it.\\
- Incorrect Answer (hand movement or object handling is slightly different but plausible): The person loosens the fastener by shaking the piece in one hand while holding the wrench in the other hand.\\

The sentences should:\\
- Contain same objects but different hand movements or ways of doing the action compared to the correct answer\\
- Be incompatible with the correct answer\\
- Maintain the same level of detail and be of a similar length to the correct answer\\
- Avoid overlapping with each other\\

The answer format should be json:\\
\{"1": "<sentence1>", "2": "<sentence2>", "3": "<sentence3>", ..., "4": "<sentence4>"\}\\

Make sure that you only answer json.\\
\\
Question: \textcolor{blue}{\{question\}}\\
Correct Answer: \textcolor{blue}{\{answer\}}
}
    \end{tcolorbox}
    \caption{Prompt to generate options for \textbf{Process} question. We replace \textcolor{blue}{\{question\}} and \textcolor{blue}{\{answer\}} with created question and answer, respectively}
    \label{tab:prompt_option_generation_process}
\end{table}

\begin{table}[h]
    \centering
    \begin{tcolorbox}
    \small{
    \textbf{User Prompt}:\\
    You will be given a pair of a question and an answer about a hand-object interaction.\\
Create 4 sentences that describe similar but incorrect answers, as per the examples.\\
Examples:\\
- Question: What object is used by the hands?\\
- Correct Answer: knife,apple.\\
- Incorrect Answer: chopping board.\\

The sentences should:\\
- Contain incorrect object names that are likely to be in the same scene as the correct answer (e.g., chopping board)\\
- Even if the correct answer contains multiple objects, the incorrect answer should contain only one object per option\\
- Avoid overlapping with each other\\

The answer format should be json:\\
\{"1": "<sentence1>", "2": "<sentence2>", "3": "<sentence3>", ..., "4": "<sentence4>"\}\\

Make sure that you only answer json.\\
\\
Question: \textcolor{blue}{\{question\}}\\
Correct Answer: \textcolor{blue}{\{answer\}}
}
    \end{tcolorbox}
    \caption{Prompt to generate options for \textbf{Objects} question. We replace \textcolor{blue}{\{question\}} and \textcolor{blue}{\{answer\}} with created question and answer, respectively}
    \label{tab:prompt_option_generation_object}
\end{table}

\begin{table}[h]
    \centering
    \begin{tcolorbox}
    \small{
    \textbf{User Prompt}:\\
    You will be given a pair of a question and an answer about a hand-object interaction.\\
Create 4 sentences that describe similar but incorrect answers, as per the examples.\\
Examples:\\
- Question: Where does the person place the cup?\\
- Correct Answer: On the left bottom corner of the table.\\
- Incorrect Answer: On the right top corner of the table.\\
- Incorrect Answer: On the chair under the table.\\

The sentences should:\\
- Include a specific location indicating where the object is moved to, but it should be a different location from the correct answer\\
- Be close to but different from the correct location (e.g., "On the right top corner of the table.")\\
- Maintain the same level of detail and be of a similar length to the correct answer\\
- Avoid overlapping with each other\\

The answer format should be json:\\
\{"1": "<sentence1>", "2": "<sentence2>", "3": "<sentence3>", ..., "4": "<sentence4>"\}\\

Make sure that you only answer json.\\
\\
Question: \textcolor{blue}{\{question\}}\\
Correct Answer: \textcolor{blue}{\{answer\}}
}
    \end{tcolorbox}
    \caption{Prompt to generate options for \textbf{Location} question. We replace \textcolor{blue}{\{question\}} and \textcolor{blue}{\{answer\}} with created question and answer, respectively}
    \label{tab:prompt_option_generation_location}
\end{table}

\begin{table}[h]
    \centering
    \begin{tcolorbox}
    \small{
    \textbf{User Prompt}:\\
    You will be given a pair of a question and an answer about a hand-object interaction.\\
Create 4 sentences that describe similar but incorrect answers, as per the examples.\\
Examples:\\
- Question: How did the state of the screw change?\\
- Correct Answer: The screw was picked up, put on the hole, and turned clockwise.\\
- Incorrect Answer (a little bit different state change): The screw that had already been in the hole was turned counter-clockwise.\\
- Question: How did the state of the camera change?\\
- Correct Answer: The camera was moved right by the right hand.
- Incorrect Answer (a little bit different state change): The camera was moved left and slightly shifted.\\

The sentences should:\\
- Describe very similar but different state changes of the object.\\
- Be incompatible with the correct answer.\\
- Not change the object (tools) used from the correct answer.\\
- Maintain the same level of detail and be of a similar length to the correct answer.\\
- Avoid overlapping with each other.\\

The answer format should be json:\\
\{"1": "<sentence1>", "2": "<sentence2>", "3": "<sentence3>", ..., "4": "<sentence4>"\}\\

Make sure that you only answer json.\\
\\
Question: \textcolor{blue}{\{question\}}\\
Correct Answer: \textcolor{blue}{\{answer\}}
}
    \end{tcolorbox}
    \caption{Prompt to generate options for \textbf{State} question. We replace \textcolor{blue}{\{question\}} and \textcolor{blue}{\{answer\}} with created question and answer, respectively}
    \label{tab:prompt_option_generation_state}
\end{table}

\begin{table}[h]
    \centering
    \begin{tcolorbox}
    \small{
    \textbf{User Prompt}:\\
    You will be given a pair of a question and an answer about a hand-object interaction.\\
Create 4 sentences that describe similar but incorrect answers, as per the examples.\\
Examples:\\
- Question: What part of the hammer is being held?\\
- Correct Answer: The bottom of the hammer handle.\\
- Incorrect Answer: Close to the head of the hammer.\\

The sentences should:\\
- Contain nonexistent parts or incorrect parts of the object (e.g., Close to the head of the hammer)\\
- Maintain the same level of detail and be of a similar length to the correct answer\\
- Avoid overlapping with each other\\

The answer format should be json:\\
\{"1": "<sentence1>", "2": "<sentence2>", "3": "<sentence3>", ..., "4": "<sentence4>"\}\\

Make sure that you only answer json.\\
\\
Question: \textcolor{blue}{\{question\}}\\
Correct Answer: \textcolor{blue}{\{answer\}}
}
    \end{tcolorbox}
    \caption{Prompt to generate options for \textbf{Parts} question. We replace \textcolor{blue}{\{question\}} and \textcolor{blue}{\{answer\}} with created question and answer, respectively}
    \label{tab:prompt_option_generation_part}
\end{table}

\FloatBarrier

\begin{table*}[ht]
\centering
\small
\begin{tabular}{p{4cm} p{10cm}}
\hline
\textbf{Purpose} & To ask how the person is manipulating objects and to understand the dynamically changing relationships between the hands and the objects. \\
\textbf{Required answer information} & 
\begin{itemize}[leftmargin=1.5em]
  \item Details of manipulation by both hands (left and right, if distinguishable)
  \item All objects involved in the action
\end{itemize} \\
\textbf{Example question} & What is the person doing with his/her hands? \\
\textbf{Example correct answer} & The person is slicing an apple on the chopping board using the knife in his right hand while holding the apple with his left hand. \\
\textbf{Examples of incorrect answers} & 
\begin{itemize}[leftmargin=1.5em]
  \item The person is slicing an apple on the table using an apple slicer in his right hand while holding the apple with his left hand.
  \item The person is removing the skin of an apple on the chopping board using the knife in his right hand while grasping the apple with his left hand.
\end{itemize} \\
\textbf{Notes} & Do not omit the subject (“The person”). \\
\hline
\end{tabular}
\caption{Instruction for annotators to annotate \textbf{Action} category}
\label{tab:instruction_action}
\end{table*}

\begin{table*}[ht]
\centering
\small
\begin{tabular}{p{4cm} p{10cm}}
\hline
\textbf{Purpose} & To ask about the manner, procedure, technique, or skill involved in a hand action or its interaction with an object. \\
\textbf{Required answer information} & 
\begin{itemize}[leftmargin=1.5em]
  \item Which hand is used
  \item How the hand moves or interacts with the object, including the steps and state changes
\end{itemize} \\
\textbf{Example question} & How does the person drop the toy on the table? \\
\textbf{Example correct answer} & The person drops the action figure gently on the table by holding it with the index finger and thumb in the left hand. \\
\textbf{Incorrect example} & The person released his grip and let it fall with force. \\
\textbf{Notes} & Do not omit the subject (“The person”) or the verb (action). \\
\hline
\end{tabular}
\caption{Instruction for annotators to annotate \textbf{Process} category}
\label{tab:instruction_process}
\end{table*}

\begin{table*}[ht]
\centering
\small
\begin{tabular}{p{4cm} p{10cm}}
\hline
\textbf{Purpose} & To identify the types and positions of objects being manipulated by the hands. \\
\textbf{Required answer information} & 
\begin{itemize}[leftmargin=1.5em]
  \item A verbal description of all manipulated objects and their positions (including segmentation mask if applicable)
  \item Objects that are merely touched and not clearly manipulated should not be included as correct or incorrect options
\end{itemize} \\
\textbf{Example question} & What object is used by the hands? \\
\textbf{Correct answer} & knife, apple \\
\textbf{Incorrect example} & The chopping board (present but not manipulated) \\
\textbf{Notes} & Only include object names in the answer. Separate multiple correct objects with commas. \\
\hline
\end{tabular}
\caption{Instruction for annotators to annotate \textbf{Objects} category}
\label{tab:instruction_objects}
\end{table*}

\begin{table*}[ht]
\centering
\small
\begin{tabular}{p{4cm} p{10cm}}
\hline
\textbf{Purpose} & To ask where the manipulated object was moved to, or where it ended up as a result of the action. \\
\textbf{Required answer information} & A specific description of the location where the object was placed or moved. \\
\textbf{Example question} & Where does the person place the cup? \\
\textbf{Example correct answer} & The person places the cup on the left bottom corner of the table. \\
\textbf{Incorrect example} & The person places the cup on the top right corner of the table. \\
\textbf{Notes} & Do not omit the subject (“The person”), the verb (action), or the object. \\
\hline
\end{tabular}
\caption{Instruction for annotators to annotate \textbf{Location} category}
\label{tab:instruction_location}
\end{table*}

\begin{table*}[ht]
\centering
\small
\begin{tabular}{p{4cm} p{10cm}}
\hline
\textbf{Purpose} & To ask how the state, structure, composition, or spatial arrangement of the object changed during the video (or remained unchanged). \\
\textbf{Required answer information} & A description of how the object’s state, structure, composition, or placement changed or did not change in the video. \\
\textbf{Example question} & How did the state of the apple change? \\
\textbf{Correct answer} & The apple was cut into small slices. \\
\textbf{Example question 2} & How did the state of the camera change? \\
\textbf{Correct answer 2} & The camera is divided into two parts: the body and the lens. \\
\textbf{Incorrect examples} & 
\begin{itemize}[leftmargin=1.5em]
  \item The apple was crushed.
  \item The apple was sliced. (when slicing did not occur)
\end{itemize} \\
\textbf{Notes} & Do not omit the subject (the object) or the verb. \\
\hline
\end{tabular}
\caption{Instruction for annotators to annotate \textbf{State} category}
\label{tab:instruction_state}
\end{table*}

\begin{table*}[ht]
\centering
\small
\begin{tabular}{p{4cm} p{10cm}}
\hline
\textbf{Purpose} & To identify the specific part of the object that is being affected, considering the object’s structure, function, and spatial relation to the hands. \\
\textbf{Required answer information} & A detailed verbal description of the affected region and its position (including segmentation mask if applicable). \\
\textbf{Example question} & What part of the hammer is being held? \\
\textbf{Correct answer} & The bottom of the hammer handle is being held. \\
\textbf{Incorrect example} & Close to the head of the hammer is being held. \\
\textbf{Notes} & Do not omit the subject (the part) or the verb (effect). \\
\hline
\end{tabular}
\caption{Instruction for annotators to annotate \textbf{Parts} category}
\label{tab:instruction_parts}
\end{table*}

\clearpage

\paragraph{Final QA post-processing using LLMs.}
After human QA annotation, we refined each option, including the correct answer, using LLMs to correct their grammar and ensure a consistent tone across all choices, without changing their meaning. This was especially done for the \textbf{Action}, \textbf{Process}, and \textbf{State} categories, where longer sentences tend to introduce textual biases (e.g., the correct answer being more likely to contain grammatical errors than the distractors). All the prompts used for each question category are shown in Table~\ref{tab:prompt_correct_action} and Table~\ref{tab:prompt_correct_state}.

\begin{table}[h]
    \centering
    \begin{tcolorbox}
    \small{
    You will be given a triplet consisting of a question, a correct answer, and a set of distractors.\\
First, revise the answer if it includes grammatical errors or is not in a natural tone without changing its meaning. If the answer is already correct, please keep it as is.\\
Then, rephrase each distractor to make it more plausible and similar in tone to the correct answer, without significantly changing its original meaning, since the distractors are currently written in a biased way, making them too easy to eliminate.\\
To do this, you may:\\
- Use words or phrasing that commonly appear in the correct answer, or avoid words frequently used in distractors.\\
- Tone down any exaggeration to make the distractors sound more natural and believable.\\
- Remove or rephrase strong negations (such as “without” or “instead”) if they clearly oppose the correct answer, unless they are essential to the meaning.\\
\\
The answer (rephrased distractors) format should be json:\\
\{\\
    "answer": "<revised\_answer>",\\
    "options":\{"1": "<sentence1>", "2": "<sentence2>", "3": "<sentence>", ..., "4": "<sentence4>" \}\\
\}\\
\\
Make sure that you only answer json.\\
Note that all the sentences should start with in the same way as the original sentences (mostly "The person...").\\
\\
Question: \textcolor{blue}{\{question\}}\\
Correct Answer: \textcolor{blue}{\{answer\}}\\
Distractors: \textcolor{blue}{\{options\}}\\
}
    \end{tcolorbox}
    \caption{Prompt to refine options for \textbf{Action} and \textbf{Process} question. We replace \textcolor{blue}{\{question\}}, \textcolor{blue}{\{answer\}}, \textcolor{blue}{\{options\}} with created question, answer, and options respectively}
    \label{tab:prompt_correct_action}
\end{table}

\begin{table}[h]
    \centering
    \begin{tcolorbox}
    \small{
    You will be given a triplet consisting of a question, a correct answer, and a set of distractors.\\
First, revise the answer if it includes grammatical errors or is not in a natural tone without changing its meaning.\\
Note that since the answer is mainly written in the passive voice to focus on the state of the object, please make sure to keep it in passive form. If the answer is already correct, please keep it as is.\\
\\
Then, rephrase each distractor to make it more plausible and similar in tone to the correct answer, without significantly changing its original meaning, since the distractors are currently written in a biased way, making them too easy to eliminate.
Also, it's better to avoid using adverbs (e.g., "gently") in the distractors, since adverbs typically describe human action rather than the state change of the object.\\
\\
The answer (rephrased distractors) format should be json:\\
\{\\
    "answer": "<revised\_answer>",\\
    "options":\{"1": "<sentence1>", "2": "<sentence2>", "3": "<sentence>", ..., "4": "<sentence4>" \}\\
\}\\
\\
Make sure that you only answer json.\\
Note that all the sentences should start with in the same way as the original sentences (mostly "The person...").\\
\\
Question: \textcolor{blue}{\{question\}}\\
Correct Answer: \textcolor{blue}{\{answer\}}\\
Distractors: \textcolor{blue}{\{options\}}\\
}
    \end{tcolorbox}
    \caption{Prompt to refine options for \textbf{State} question. We replace \textcolor{blue}{\{question\}}, \textcolor{blue}{\{answer\}}, \textcolor{blue}{\{options\}} with created question, answer, and options respectively}
    \label{tab:prompt_correct_state}
\end{table}

\paragraph{Choice of LLMs.}

We used \texttt{gpt-4o-mini-2024-07-18} to generate the initial QA pairs. For final refinement, we used \texttt{gpt-4o-2024-08-06}.
We note that generated questions/options are for reference and all the pairs were thoroughly reviewed and corrected to form the final QA pairs.

\paragraph{Human mask annotation.}
Egocentric videos often include severe blurring that harms the visual quality of the video clip.
To this end, we opted to annotate three representative frames from each 5-second video clip.
Annotators manually selected one frame each from the front, middle, and last thirds of the video clip, where the target regions were clearly visible and sampled them from different parts of the video whenever possible.
However, the above condition was relaxed when the frames from some of the intervals were unusable.

For \textbf{Object Parts} questions, clips often involve object state changes, which change the appearance of the components.
In such cases, both the frames before and after the state change were selected.

\paragraph{Human Evaluation.}
We recruited three human raters who each passed the initial screening, achieving over 70\% accuracy on a set of 30 questions sampled from the validation split of our dataset.
Table~\ref{tab:human_evaluation} reports the performance of the raters and their agreement ratio. Agreement denotes the proportion of questions for which all three raters selected exactly the same set of options. The \textbf{Objects} category shows relatively lower agreement because multiple answers may be correct, reducing the likelihood that all raters choose the same set of options.
Overall, humans performed over 90\% accuracy/AP across all the categories, indicating that our benchmark is solvable by humans and that the current models still have a significant gap between human performance.

\begin{table}[t]
\centering
\resizebox{\linewidth}{!}{
\begin{tabular}{lccccccc}
\toprule
\multirow{2}{*}{\textbf{Human}} & \cellcolor{actColor} \textbf{Action} & \cellcolor{proColor} \textbf{Process} & \cellcolor{objColor} \textbf{Objects}  & \cellcolor{locColor} \textbf{Location} & \cellcolor{staColor} \textbf{State} & \cellcolor{parColor} \textbf{Parts} & \textbf{Avg.} \\

& (Acc) & (Acc) & (AP) & (Acc) & (Acc) & (Acc) & (Acc)\\

\midrule
Rater 1 & 98.9 & 97.6 & 96.6 & 94.2 & 99.0 & 99.1 & 97.7 \\
Rater 2 & 99.0 & 98.1 & 94.2 & 97.0 & 91.3 & 97.4 & 96.5 \\
Rater 3 & 97.9 & 92.0 & 97.1 & 98.7 & 95.7 & 94.2 & 95.7 \\
\midrule
\textbf{Avg.} & 98.6 & 95.9 & 96.0 & 96.6 & 95.3 & 96.9 & 96.6 \\
\midrule
\textbf{Agreement} & 95.8 & 87.8 & 76.5 & 90.1 & 87.0 & 90.7 & 90.3 \\
\bottomrule
\end{tabular}
}
\caption{\textbf{Human evaluation results.} 
Each rater independently selected correct answer(s) for each question. 
\textbf{Agreement} indicates proportion of questions for which all three raters selected exactly same set of options.}
\label{tab:human_evaluation}
\end{table}

\paragraph{Detailed dataset statistics.}
\begin{figure*}[t]
    \centering
    \includegraphics[width=1\linewidth]{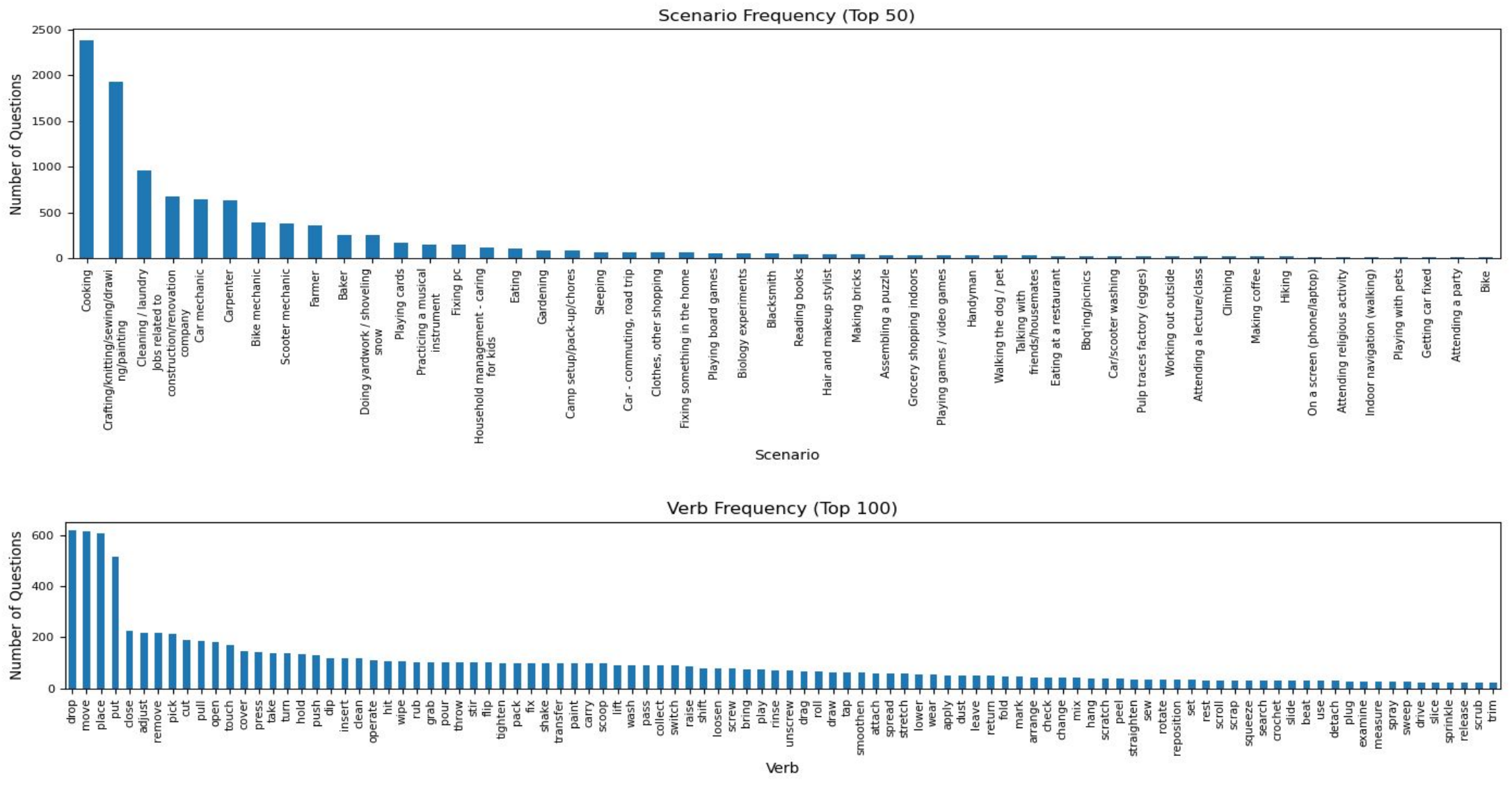}
    \caption{Top-50 scenario distribution (top) and top-100 verb frequency distribution (bottom) of \datasetnamenospace.}
    \label{fig:diversity_full}
\end{figure*}

\begin{figure*}
    \centering
    \includegraphics[width=.95\linewidth]{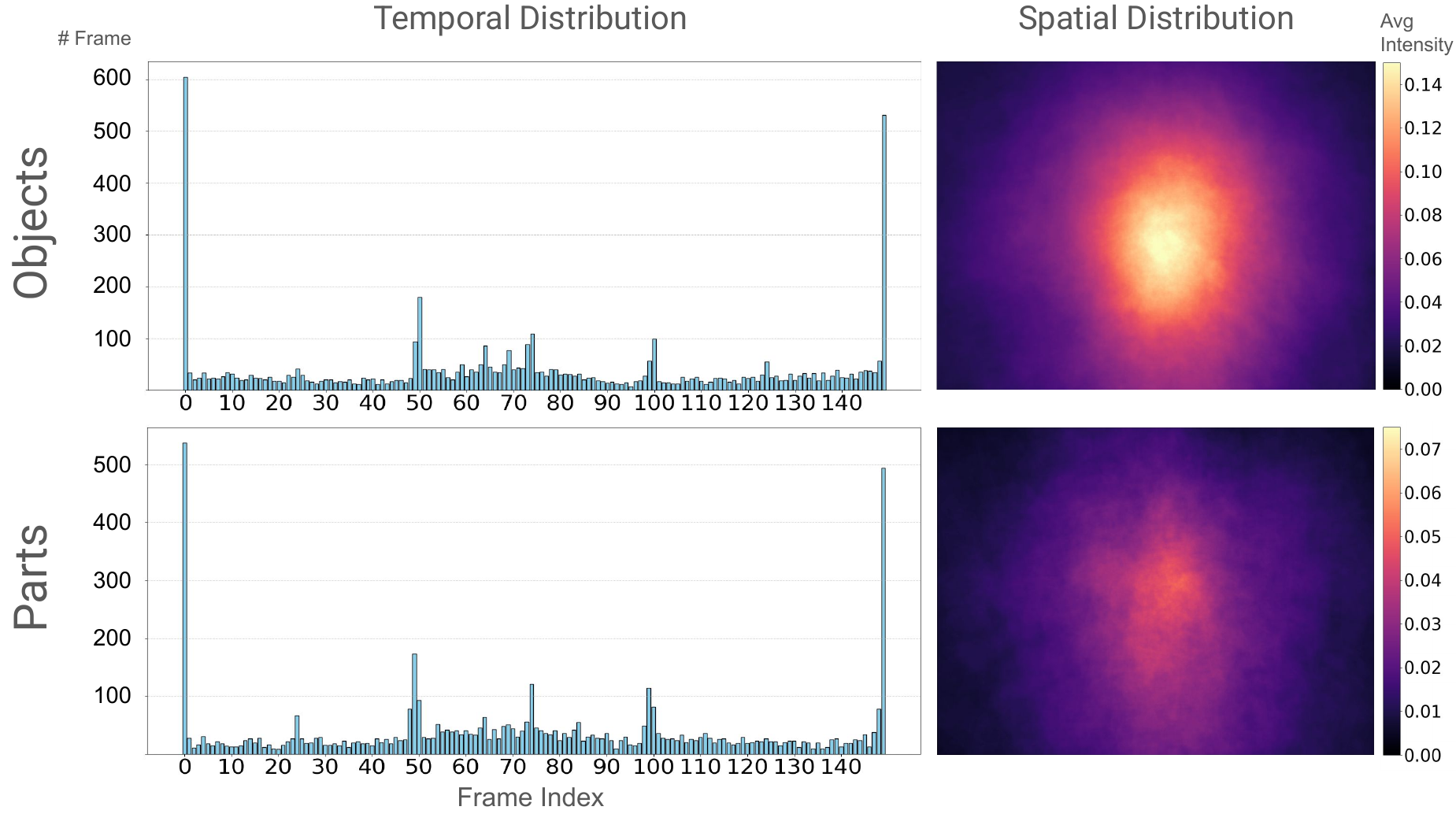}
    \caption{Temporal distribution (right) and spatial distribution (left) of mask annotations per category.}
    \label{fig:mask_spatio_temporal_distribution}
\end{figure*}

Figure~\ref{fig:diversity_full} shows the detailed distribution of the scenarios and the primary verbs included in \datasetnamenospace.
While the scenarios reflect the distribution of the original Ego4D video clips, the verbs are more uniformly selected to ensure diverse HOIs are covered.

Figure~\ref{fig:mask_spatio_temporal_distribution} shows the spatial and temporal distribution of segmentation mask annotations for \textbf{Objects} and \textbf{Object Parts} questions. 
While selected frames are biased towards the beginning and the end of a video, the remaining annotations are evenly distributed throughout the video.
Regarding the spatial direction, the segmentation showed a tendency to concentrate in the center of a frame, but also appeared to spread around the center.

\paragraph{Details on compensation.}
We outsourced the annotation of MCQs and segmentation masks to an agency at a total cost of 3,180,000 JPY (approximately 22,000 USD).
For the human evaluation, we commissioned a different company and hired separate personnel as raters, incurring an additional cost of 4,510,585 JPY (approximately 28,600 USD).
The company is responsible for managing payments to annotators, ensuring compliance with the minimum wage regulations in the annotator's country.

\section{Details on experimental settings}
\label{sec:details_on_experimental_settings}

\subsection{Multi-Choice Questions}
\paragraph{Frame sampling strategy.}
We sample $n$ frames uniformly from a video of length $L$ by dividing it into $n$ equal segments and selecting one frame from the center of each segment.
Specifically, the sampling index $i_k$ for the $k$-th frame ($k = 0, 1, \dots, n{-}1$) is computed as:
\[
i_k = \left\lfloor k \cdot \text{gap} + \frac{\text{gap}}{2} \right\rfloor,\quad \text{where } \text{gap} = \frac{L}{n}
\]
This approach ensures that the sampled frames are evenly distributed over the entire video, while avoiding bias toward the start or end. By choosing the center of each interval, we obtain a more representative snapshot of the temporal progression.

\paragraph{Prompts for zero-shot evaluation.}
The textual prompts fed to video-language models with integrated LLMs to solve MCQs are shown in Table~\ref{tab:prompt_zero-shot_single} and Table~\ref{tab:prompt_zero-shot_multiple}.

\paragraph{Computational cost.}
The 7B-size video-language models fit on a single NVIDIA H200 (141GB) GPU and completed inference for each question category in under an hour. The 72B-size models were able to run on a single node with eight H200 GPUs, requiring approximately 1.7 hours per category.

\subsection{Referring Video Object Segmentation}
\paragraph{Frame sampling strategy.}
Given a set of annotated frame indices $\mathcal{A} \subseteq [0, n]$ and a target number of samples $l$, we construct a set $\mathcal{S}$ of $l$ indices that are both representative and temporally balanced.

\begin{itemize}
    \item We initialize the set $\mathcal{S}$ as the sorted, unique subset of annotated indices $\mathcal{A}$ within their valid range:
    \[
    \mathcal{S} \gets \mathrm{sorted}\left(\{ x \in \mathcal{A} \mid 0 \leq x \leq n \}\right)
    \]
    \item While $|\mathcal{S}| < l$, we iteratively identify the largest temporal gap between consecutive elements in $\mathcal{S}$, including gaps at the start ($[0, \mathcal{S}_1]$) and end ($[\mathcal{S}_{|\mathcal{S}|}, n]$), and insert the midpoint of the largest such gap:
    \[
    \text{midpoint} = \left\lfloor \frac{i + j}{2} \right\rfloor \quad \text{for each gap } [i, j]
    \]
    \item This process continues until $|\mathcal{S}| = l$, or no more meaningful midpoints can be added.
    \item If the final size $|\mathcal{S}| > l$ (e.g., due to duplicate insertions), we resample $l$ indices from $\mathcal{S}$ to evenly cover $[0, n]$. Specifically, we define $l$ ideal positions:
    \[
    t_i = \mathrm{round}\left(\frac{i \cdot n}{l - 1}\right), \quad i = 0, 1, \dots, l{-}1
    \]
    and for each $t_i$, we select the closest available index in $\mathcal{S}$ without duplication.
\end{itemize}

This strategy ensures that manually annotated indices are selected while interpolating additional indices to maximize temporal coverage.

\paragraph{Prompts for zero-shot evaluation.}
The prompt used for the baseline that takes questions as input is provided in Table~\ref{tab:prompt_zero-shot_seg}.
The prompt that uses the ground-truth option is provided in Table~\ref{tab:prompt_gt_seg}.

\paragraph{Grouping for evaluation.}
We group videos into three size bins—small (S), medium (M), and large (L)—based on the average area of their ground-truth masks, so we can examine performance as a function of target size. The S/M/L thresholds differ between the \textbf{Area} and \textbf{Object} settings:
\begin{itemize}
    \item \textbf{Area}: S→M at 372, M→L at 2,127 (pixels)
    \item \textbf{Object}: S→M at 3,581, M→L at 13,063 (pixels)
\end{itemize}

\paragraph{Computational cost.}
Inference with the VideoLISA-3.8B model was performed on a single NVIDIA H200 GPU and took roughly 1.5 hours per category.
Inference with the Sa2VA-8B model on a single NVIDIA H200 GPU required about 2 hours per category for the video baseline and around 1.5 hours for the frame-wise baseline.

\subsection{Integration of HOI cues}
\label{subsec:details_on_hoi_cues}
\begin{figure}
    \centering
    \includegraphics[width=1\linewidth]{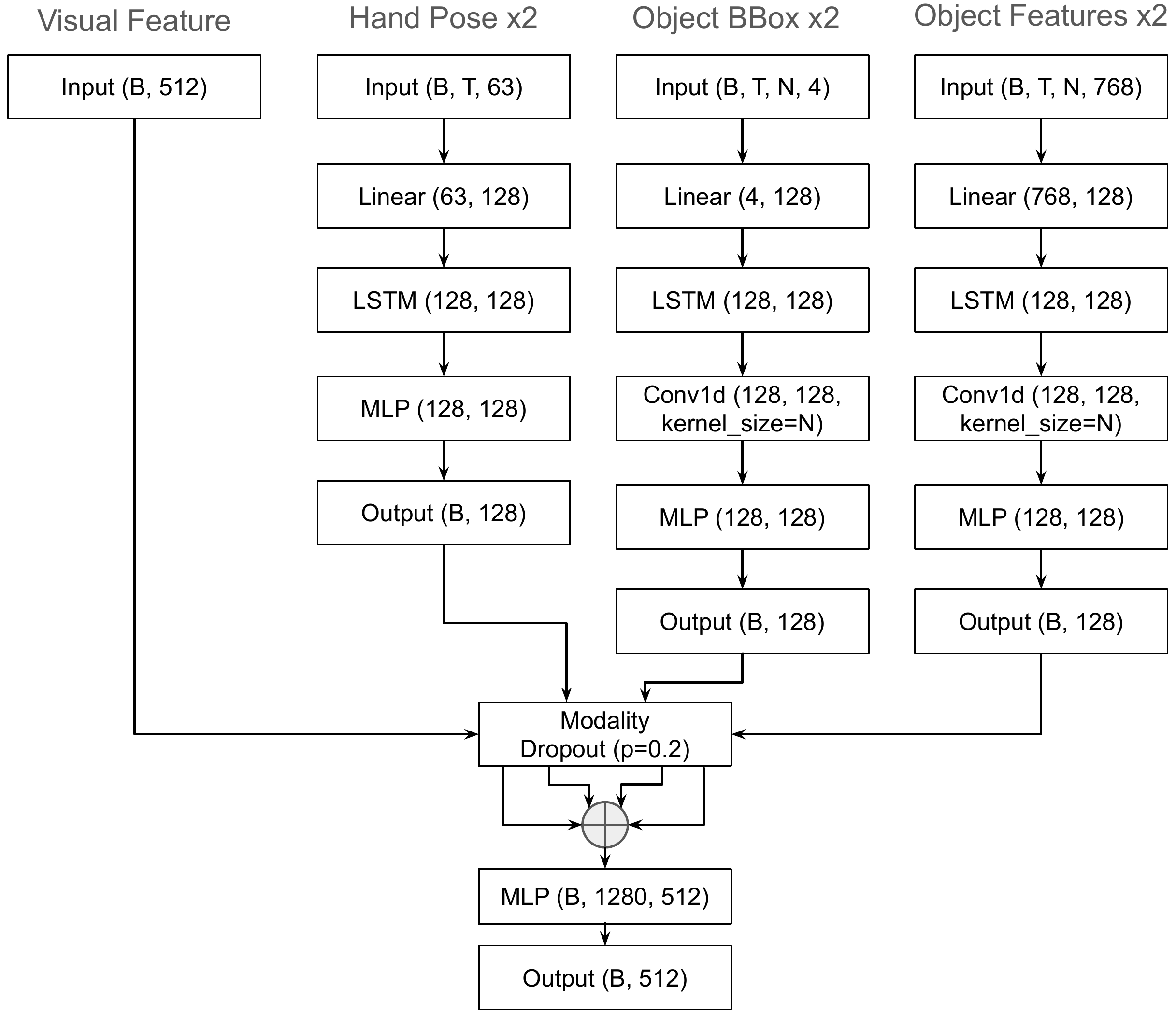}
    \caption{Architecture of additional head used for fine-tuning experiment. Branches are omitted for unused modalities.}
    \label{fig:hoi_cue_model}
\end{figure}

\paragraph{Implementation details.}
Figure~\ref{fig:hoi_cue_model} illustrates the architecture of our model, which integrates multiple modalities: frame-level RGB visual features, hand poses, bounding boxes of manipulated objects, and visual features of the manipulated objects.

Hand poses are extracted using an off-the-shelf 3D hand pose detector~\cite{potamias2024wilor}, resulting in a tensor of shape $[B, T, 21, 3]$ for each hand, where $B$ is the batch size and $T$ is the number of frames in the video. The bounding boxes of manipulated objects for each hand are obtained using AMEGO~\cite{goletto2024amego} and represented as $[B, T, N, 4]$, where $N$ is the maximum number of objects per hand. We set $N=8$ in our experiments. For each detected bounding box, we crop the corresponding image region and extract CLIP features~\cite{radford2021learning}. 

For each modality, we use separate processing modules for the left and right sides, resulting in seven feature vectors in total. All modality-specific features, except for the global RGB visual feature, have a shape of $[B, 128]$.
Before concatenation, we apply a modality-dropout layer in which entire modalities (i.e., both left- and right-hand streams of a modality) may be jointly dropped. Up to two such modalities are randomly dropped with a probability of $p=0.2$, encouraging the model to rely on the remaining modalities.
Finally, all features are concatenated and passed through a multi-layer perceptron (MLP) to produce the final fused representation of shape $[B, 512]$.

\paragraph{Training.}
For each integrated feature $\mathbf{v}_i$, we associate one positive text 
feature and $B_N$ negative text features sampled from (i) distractors within the 
same video and (ii) answer options of other videos in the batch. The contrastive 
training is summarized in Algorithm~\ref{alg:contrastive}.

\begin{algorithm}[t]
\caption{Contrastive Training for Integrated Features}
\label{alg:contrastive}
\begin{algorithmic}[1]
\Require Integrated feature $\mathbf{v}_i$, positive text feature $\mathbf{p}_i$, 
         negative features $\{\mathbf{n}_{i,j}\}_{j=1}^{B_N}$, temperature $\tau$
\State $\hat{\mathbf{v}}_i \gets \mathrm{normalize}(\mathbf{v}_i)$
\State $\hat{\mathbf{p}}_i \gets \mathrm{normalize}(\mathbf{p}_i)$
\For{$j = 1$ to $B_N$}
    \State $\hat{\mathbf{n}}_{i,j} \gets \mathrm{normalize}(\mathbf{n}_{i,j})$
\EndFor
\State Compute logits:
\[
\mathbf{s}_i = 
\left[
\frac{\hat{\mathbf{v}}_i^\top \hat{\mathbf{p}}_i}{\tau},\ 
\frac{\hat{\mathbf{v}}_i^\top \hat{\mathbf{n}}_{i,1}}{\tau},\ 
\dots,\ 
\frac{\hat{\mathbf{v}}_i^\top \hat{\mathbf{n}}_{i,B_N}}{\tau}
\right]
\]
\State $\mathbf{y}_i \gets [1, 0, \dots, 0]$
\State $\mathcal{L}_i \gets \mathrm{BCEWithLogits}(\mathbf{s}_i, \mathbf{y}_i)$
\end{algorithmic}
\end{algorithm}

The total loss is computed by averaging across all samples in the batch:
\[
\mathcal{L} = \frac{1}{B} \sum_{i=1}^{B} \mathcal{L}_i.
\]




We trained separate models for each category. The batch size was set to 16, the learning rate to $1 \times 10^{-5}$, the weight decay to $1 \times 10^{-4}$, the number of negative samples $B_N$ to 16, and the temperature $\tau$ to 0.07. Each model was trained for 500 epochs, and the best-performing checkpoint was selected based on validation performance.

\paragraph{Computational cost.}
As shown in Table~\ref{tab:cost_additional_layers}, the additional branches for HOI cues introduce only minor computational overhead. Table~\ref{tab:cost_feature_extraction} further reports the cost of extracting each HOI cue.
For training, all the model variants fit on a single NVIDIA H200 GPU and finish in roughly 0.05--2.5 hours per category, depending on the number of additional cues.

\begin{table}[t]
    \centering
    \resizebox{\linewidth}{!}{
    \begin{tabular}{lcc}
        \toprule
        {} & \textbf{Extra Params} & \textbf{Extra GFLOPs / video} \\
        \midrule
        Hand Pose Branch & 0.80M & 0.0052 \\
        Object BBox Branch & 0.92M & 0.0351 \\
        Object Feat Branch & 1.02M & 0.0602 \\
        \bottomrule
    \end{tabular}
    }
    \caption{Computational cost of additional layers for HOI cues.}
    \label{tab:cost_additional_layers}
\end{table}

\begin{table}[t]
    \centering
    \resizebox{\linewidth}{!}{
    \begin{tabular}{lcc}
        \toprule
        {} & \textbf{Params} & \textbf{GFLOPs / frame} \\
        \midrule
        Hand Pose (YOLOv8 det.\ + WiLoR pose) & 26.63M + 693.03M & 41.56 + 140.09 \\
        Object BBox (Faster R-CNN, ResNet-101) & 47.36M & 219.57 \\
        Object Feats (CLIP) & 202.05M & 51.90 \\
        \bottomrule
    \end{tabular}
    }
    \caption{Computational cost of feature extraction for HOI cues.}
    \label{tab:cost_feature_extraction}
\end{table}

\begin{table}[h]
    \centering
    \begin{tcolorbox}
    \small{
    Carefully watch the first-person view video and pay attention to the cause and sequence of events, the details and movements of objects, and the actions and poses of persons.\\
    Question: \textcolor{blue}{\{question\}}\\
    Choose **only one** option from the following list.\\
    Options:\\
    (A) \textcolor{blue}{\{option1\}}\\
    (B) \textcolor{blue}{\{option2\}}\\
    (C) \textcolor{blue}{\{option3\}}\\
    (D) \textcolor{blue}{\{option4\}}\\
    (E) \textcolor{blue}{\{option5\}}\\
    
    Answer format:\\
    (A) <Description of Option A>
}
    \end{tcolorbox}
    \caption{Prompt for zero-shot evaluation of video-language models integrated with LLMs when there is only one correct answer. \textcolor{blue}{\{question\}} is replaced with question and \textcolor{blue}{\{option$n$\}} is replaced with $n$-th option.}
    \label{tab:prompt_zero-shot_single}
\end{table}

\begin{table}[h]
    \centering
    \begin{tcolorbox}
    \small{
    Carefully watch the first-person view video and pay attention to the cause and sequence of events, the detail and movement of objects, and the action and pose of persons.\\
    Question: \textcolor{blue}{\{question\}}\\
    Choose **all** options that apply from the following list.\\
    Options:\\
    (A) \textcolor{blue}{\{option1\}}\\
    (B) \textcolor{blue}{\{option2\}}\\
    (C) \textcolor{blue}{\{option3\}}\\
    (D) \textcolor{blue}{\{option4\}}\\
    (E) \textcolor{blue}{\{option5\}}\\
    ...\\
    
    Answer format:\\
    (A) <Description of Option A>
}
    \end{tcolorbox}
    \caption{Prompt for zero-shot evaluation of video-language models integrated with LLMs when there are multiple correct answers. \textcolor{blue}{\{question\}} is replaced with question and \textcolor{blue}{\{option$n$\}} is replaced with $n$-th option.}
    \label{tab:prompt_zero-shot_multiple}
\end{table}

\begin{table}[h]
    \centering
    \begin{tcolorbox}
    \small{Segment the area that corresponds to the answer to the question.\\
    Question:  \textcolor{blue}{\{question\}}
}
    \end{tcolorbox}
    \caption{Prompt for Sa2VA (frame-wise/video) baseline in referring video object segmentation. \textcolor{blue}{\{question\}} is replaced with question.}
    \label{tab:prompt_zero-shot_seg}
\end{table}

\begin{table}[h]
    \centering
    \begin{tcolorbox}
    \small{Segment all the mentioned area:\\
    \textcolor{blue}{\{GT\}}
}
    \end{tcolorbox}
    \caption{Prompt for GT + Sa2VA baseline in referring video object segmentation. \textcolor{blue}{\{GT\}} is replaced with ground truth.}
    \label{tab:prompt_gt_seg}
\end{table}

\clearpage
\section{Fine-tuning Qwen2.5-VL-7B}
\label{sec:fine-tune_qwen}
We conducted instruction tuning on the training split of \datasetname~to explore the effectiveness of fine-tuning. Specifically, we used Qwen2.5-VL-7B as the base model and trained LoRA adapters using QLoRA~\cite{dettmers2023qlora}.

\paragraph{Implementation details.}
We trained separate models for each category. We only trained the LoRA parameters injected into the query and value projection layers (q\_proj and v\_proj) of the attention modules, while keeping all other model weights frozen. The number of input video frames is $16$, and the resolution is $224 \times 398$. We set the batch size to 4, learning rate to $1 \times 10^{-4}$. Each model was trained for 150 epochs, and the best-performing checkpoint was selected based on the validation performance.

\paragraph{Computational cost.}
Training was performed on a single NVIDIA H200 GPU and took roughly two hours per category.

\paragraph{Results.}
Table~\ref{tab:fine-tuning_qwen} shows the results of fine-tuning the model. The \textbf{Process} and \textbf{Location} categories show improvements of roughly 10 points, followed by a 6-point gain in the \textbf{State} category, while gains in \textbf{Objects} and \textbf{Parts} remain below 5 points. A slight performance drop is observed for the \textbf{Action} category. These results indicate that small-scale instruction tuning offers limited benefits for relatively easier conventional tasks such as \textbf{Action} and \textbf{Objects}, but yields larger gains for categories that involve longer textual descriptions, likely because LLMs are better at leveraging textual biases.

\begin{table}[t]
    \centering
    \resizebox{\linewidth}{!}{%
\begin{tabular}{l|cccccc|c}
    \toprule
    \textbf{Models} & \cellcolor{actColor} \textbf{Action} & \cellcolor{proColor} \textbf{Process} & \cellcolor{objColor} \textbf{Objects}  & \cellcolor{locColor} \textbf{Location} & \cellcolor{staColor} \textbf{State} & \cellcolor{parColor} \textbf{Parts} & \textbf{Avg.} \\
      & (Acc) & (Acc) & (AP) & (Acc) & (Acc) & (Acc) & (Acc)  \\
    \midrule
    \rowcolor{gray!20}
    Qwen2.5-VL-7B zero-shot & \textbf{58.6} & 53.4 & 53.5 & 45.5 & 57.1 & 47.8 & 52.5 \\
    Qwen2.5-VL-7B fine-tuned & 58.2 & \textbf{64.9} & \textbf{57.5} & \textbf{64.8} & \textbf{63.1} & \textbf{50.2} & \textbf{60.2} \\
    \bottomrule
\end{tabular}
    }
    \caption{Results of fine-tuning Qwen2.5-VL-7B model on HanDyVQA.}
    \label{tab:fine-tuning_qwen}
\end{table}

\section{Broader Impacts}
\label{sec:broader_impacts}

The proposed \datasetname~dataset provides a detailed evaluation of fine-grained hand-object interactions. As such, it serves as a valuable benchmark for systems designed to assist human workers using visual information captured by wearable cameras in diverse real-world scenarios~\cite{plizzari2024outlook}. This enables the development of systems that can better understand subtle interactions and deliver more accurate and context-aware feedback to the users.

Such recognition capabilities are also essential for applications in Augmented Reality (AR) and Virtual Reality (VR), where systems must respond to users actions and changes in the environment in real time. Unlike previous datasets that focus primarily on action recognition or object detection, \datasetname\  offers a unique benchmark that evaluates a model’s ability to comprehend nuanced hand-object interactions and underlying dynamics, pushing the boundaries beyond conventional video recognition tasks.





\end{document}